\documentclass[5p,times]{elsarticle}

\usepackage{url}
\usepackage{amsmath}
\usepackage{graphicx}
\usepackage{stmaryrd}
\usepackage{amssymb}
\usepackage{color}
\usepackage{multirow}
\usepackage{lineno}
\usepackage{algorithm}
\usepackage{algorithmic}
\usepackage{lscape} 
\usepackage{subcaption}
\usepackage{changepage}

\usepackage{scrextend}

\begin{document}

\begin{frontmatter}

\title{A deep learning guided memetic framework for graph coloring problems}

\author[Angers]{Olivier Goudet}
\ead{olivier.goudet@univ-angers.fr}
\author[Angers]{Cyril Grelier}
\ead{Cyril.Grelier@univ-angers.fr}
\author[Angers]{Jin-Kao Hao \corref{cor1}}
\ead{jin-kao.hao@univ-angers.fr}
\cortext[cor1]{Corresponding author.} 
\affiliation[Angers]{organization={LERIA, Université d'Angers}, addressline={2 Boulevard Lavoisier, 49045 Angers}, country={France}}

\date{}

%\linenumbers

\begin{abstract}

Given an undirected graph $G=(V,E)$ with a set of vertices $V$ and a set of edges $E$, a graph coloring problem involves finding a partition of the vertices into different independent sets. In this paper we present  a new framework that combines a deep neural network with the best tools of classical heuristics for graph coloring. The proposed method is evaluated on two popular graph coloring problems (vertex coloring and weighted coloring). Computational experiments on well-known benchmark graphs show that the proposed approach is able to obtain highly competitive results for both problems. A study of the contribution of deep learning in the method highlights that it is possible to learn relevant patterns useful to obtain better solutions to graph coloring problems. 
\end{abstract}

\end{frontmatter}

\section{Introduction}\label{Introduction}

Graph coloring involves assigning colors to the vertices of a graph subject to certain constraints and optimization objective. The popular vertex coloring problem (COL) is the most representative example and can be stated as follows. Given an undirected graph $G=(V,E)$ with a set of vertices $V$ and a set of edges $E$, the COL is to color the vertices of $V$ such that two adjacent vertices receive different colors and  the number of colors used is minimized (this number is called the chromatic number of $G$, denoted by $\chi(G)$). This problem can also be seen as finding a partition of the vertex set $V$ into a minimum number of color groups (also called independent sets or color classes) such that two vertices linked by an edge belong to different color groups. In some variants of this conventional coloring problem, one aims to find a legal coloring of the graph while considering an alternative optimization objective. 

The typical search space of a graph coloring problem is composed of the partitions of vertices $V$ into $k$ color groups:

\begin{equation}
\mathcal{S} = \{ \{V_1,V_2,\dots,V_k\}: \cup_{i=1}^k V_i = V, V_i \cap V_j = \emptyset\}
\end{equation}

\noindent where $i \neq j, 1 \leq i,j \leq k, 1 \leq k \leq |V|$. This search space $\mathcal{S}$ is huge in general and finding an optimal solution $S^*$ is usually intractable unless P=NP, as most graph coloring problems are NP-hard.

Graph coloring problems have been studied very intensively in the past decades and many coloring methods have been proposed in the literature. A first category of methods are based on local search (also called neighborhood search). Starting from an initial solution typically constructed using a greedy heuristic, a local search algorithm improves the current solution by considering best moves in a given neighborhood. 
To escape local optima traps, local search algorithms usually incorporate dedicated mechanisms such tabu lists \cite{blochliger2008reactive,hertz1987using} or perturbation strategies \cite{jin2019solving,nogueira2021iterated}. However for very difficult instances of graph coloring, the single trajectory local search approach  is not powerful enough to locate very high quality solutions mainly due to its limited diversification capacity. To overcome this difficulty, hybrid algorithms have been proposed, in particular relying on the population-based memetic framework that combines local searches and crossovers \cite{MABook2012}. The memetic framework has been very successful in solving several graph coloring problems \cite{galinier1999hybrid,jin2014memetic,lu2010memetic,moalic2018variations,porumbel2010evolutionary}. These hybrid algorithms combine the benefits of local search for intensification with a population of high-quality solutions offering diversification possibilities. 

The memetic algorithms proposed in the literature for graph coloring typically use a small population with no more than 100 individuals. At each generation, one offspring solution is usually created by a crossover (or recombination) operator applied to two or more randomly selected individuals  from the population. One of the most popular crossovers used for graph coloring problems is the Greedy Partition Crossover (GPX) introduced in the hybrid evolutionary algorithm (HEA) \cite{galinier1999hybrid}. The GPX operator produces offspring by inheriting alternatively the largest color classes in the parent solutions. The resulting offspring is then improved by a local search procedure such as TabuCol \cite{hertz1987using}.

The crossover operator within a memetic algorithm enables the creation of new restarting points for the local search procedure, which are expected to be more promising and better than a pure random initialization. However, when using such mechanisms, there is usually no way of knowing in advance whether the new restarting point really indicates a promising area that is worth being further examined by the local search procedure. Indeed, sometimes, the use of a crossover can bring the search process back to an already visited region of the search space without any chance of further improvement, or to a new region far from the global optimum. Moreover, hybrid algorithms do not have a specific memory to store information about past searches and thus can hardly discover useful patterns that may exist among the solutions encountered during the search trajectories (though inheriting color classes with crossovers can be seen as some sort of "learning" of good patterns).

On the other hand, numerous algorithms have been proposed since decades from the machine learning community to leverage statistical learning methods for solving difficult combinatorial search problems (see the recent survey of \cite{bengio2020machine} on this topic). These attempts have been given a new lease of life, with the emergence of deep learning techniques for combinatorial optimization problems \cite{dai2017learning,wang2021deep}, inspired from the great success of the AlphaZero algorithm for combinatorial games \cite{silver2018general}. In particular, some recent works have applied reinforcement learning and deep learning to solve graph coloring problems \cite{huang2019coloring,lemos2019graph}. Nevertheless, these studies rarely exploit specific knowledge of the problem, which make these approaches more general but may limit their performance. Indeed, the results obtained by this type of approach are for the moment far from the results obtained by state of the art algorithms on graph coloring problems such as hybrid algorithms \cite{lu2010memetic,malaguti2008metaheuristic,moalic2018variations} and simulated annealing algorithms \cite{titiloye2011quantum}. We can mention however recent studies which try to take advantage of efficient local search algorithms and machine learning techniques  \cite{goudet2021population,zhou2016reinforcement} with promising results for graph coloring problems.

In this work, we aim to push further the integration of machine learning and combinatorial optimization, by proposing a new framework which combines deep neural networks with the best tools of "classical" metaheuristics for graph coloring, so as to solve very difficult graph coloring problems which still resist the best current methods. In order to achieve this integration, we revisit an idea proposed in \cite{boyan2000learning} twenty years ago. In \cite{boyan2000learning}, Boyan and Moore remarked that the performance of a local search procedure depends on the state from which the search starts and therefore proposed to use a regression algorithm to predict the results of a local search algorithm. Once learned, this predictive model can help to select new good starting points for the local search and to accelerate the search process. We exploit this idea with the use of modern deep learning techniques to better select promising crossovers among those possible ones in each generation of a memetic algorithm. We design a specific neural network architecture for graph coloring problems inspired by deep set networks \cite{NIPS2017_f22e4747,lucas2018mixed}, in order to make it invariant by permutation of the color classes. Furthermore, as training a deep neural network requires a large amount of data, we follow the recent work \cite{goudet2021massively} to adopt a large population $P$ ($|P|>\approx 10^4$) for the underlying memetic algorithm, whose individuals evolve in parallel in the search space. In order to learn the neural network and to compute all the local searches in parallel for all the individuals of the population, we leverage GPU (Graphic Processing Units) computation.

As a proof of concept, we apply this approach to solve the weighted vertex coloring problem (WVCP) and the  vertex coloring problem (COL). The classical COL is well known and has been studied for a long time. The WVCP has recently attracted a lot of interest in the literature \cite{Grelier22,nogueira2021iterated,prais2000reactive,sun2018adaptive,wang2020reduction}. In the WVCP, a strictly positive weight $w_v$ is associated to each vertex $v$. The goal of the problem is to find a legal coloring minimizing the global score:
\begin{equation}
    f(S) = \sum_{i=1}^k \underset{j \in V_i}{\text{max}} \ w_j
    \label{eq:score_WVCP}
\end{equation}

\noindent where $V_i$ ($1 \leq i \leq k$) is a color class including all the vertices receiving color $i$  and $\underset{j \in V_i}{\text{max}} \ w_j$ is the largest weight of color class $V_i$. 

One observes that the COL is a special case of the WVCP when the vertex weight $w_v$ is equal to one for all the vertices. In this case, minimizing the function $f$ (Eq. (\ref{eq:score_WVCP})) is equivalent to the minimization of the number of colors. In the presentation that follows, we focus on the WVCP. However, for computational assessments, we present experimental results for both the WVCP and the COL.

The WVCP has a number of practical applications in different fields such as matrix decomposition problems \cite{prais2000reactive}, batch scheduling \cite{gavranovic2000graph} and manufacturing \cite{hochbaum1997scheduling}. In addition to heuristic algorithms \cite{Grelier22,nogueira2021iterated,prais2000reactive,sun2018adaptive,wang2020reduction}, it has been addressed by exact methods \cite{cornaz2017solving,furini2012exact,malaguti2009models}. 

\section{General framework - revisiting the STAGE algorithm with deep learning and memetic algorithm}

Given a problem whose goal is to find an optimal solution with respect to a minimization objective $f$, 
the expected search outcome of a stochastic local search algorithm A can be defined as:

\begin{equation}
    \mathbb{E}[f_A(S)] = \sum_{S' \in \mathcal{S}} P(S \stackrel{\text{A}}{\longrightarrow} S') f(S')
\end{equation}

\noindent where $P(S \stackrel{\text{A}}{\longrightarrow} S')$ is the probability that the search starting from $S$ will terminate in state $S'$. $\mathbb{E}[f_A(S)]$ evaluates the potential interest of $S$ as a starting state for the algorithm A. 

The main idea of the STAGE algorithm \cite{boyan2000learning} was to approximate the expectation $\mathbb{E}[f_A(S)] $ by a regression approximation model $\hat{f}_A : \mathbb{R}^d \rightarrow \mathbb{R}$, taking as input the encoded real-valued feature vector $F(S)$ (with $d$ features) of a state $S$. This function $\hat{f}_A$ can be a linear regression model or a more complex non linear model such as a neural network. 

Starting from a first random initial solution $S$, and the function approximator $\hat{f}_A$, the STAGE algorithm evolves in three steps.
\begin{enumerate}
\item \textbf{Optimize f using A.} From $S$, it runs the local search algorithm A, producing a search trajectory that ends at a local optimum $S'$.
\item \textbf{Train $\hat{f}_A$.} For each point $S_i$ on the search trajectory, use $\{(F(S_i), f(S') \}$ as a new training pair for the function approximator.
\item  \textbf{Optimize $\hat{f}_A$ using hillclimbing.} Continuing from $S'$, perform a hillclimbing search on the learned objective function $\hat{f}_A$. This results in a new state $S$ which should be a new good starting point for A.
\end{enumerate}

We revisit this idea with an adaptation for each of the three steps of the STAGE algorithm.

\begin{enumerate}
\item First, regarding the first step, we run in parallel $p$ local searches with algorithm A starting from different states to generate $p$ different search trajectories (instead of a single one). This makes it possible to build a training dataset with a high diversity of examples. 
\item Secondly, regarding the step 2, we do not use any prior mapping $F$ from states to features. Following the current trend in deep learning, the embedding of the state can be directly learned in an end-to-end pipeline with a deep neural network, denoted as $f_{\theta}: \mathcal{S} \rightarrow \mathbb{R}$. We make this neural network invariant by permutation of the group of colors $V_j$ in the coloring $S \in \mathcal{S}$, which is a very important feature of all graph coloring problems, by adapting the deep set network architecture proposed in \cite{NIPS2017_f22e4747} (see Section \ref{sec:DeepLearning}). 
\item Thirdly, in the step 3 of the original STAGE algorithm,  a hillclimbing algorithm is used to optimize the current solution guided by the objective function $\hat{f}_A$.  However, as we address a very complex problem and we use a complex non-convex and non-linear function (deep neural network), it is difficult to optimize it using a hillclimbing algorithm. We tried to use more complex algorithms such as tabu search to optimize it, but there is a deeper problem, which is the question of generalization. Indeed,  if a state $S$ is too different from the states already seen before in the training dataset, and in particular if the color groups $V_j$ that composed it are too different from the color groups already seen before by the neural network, we expect that $f_{\theta}(S)$ can be very inaccurate for the estimation of  $\mathbb{E}[f_A(S)]$. Therefore, we propose to replace this hillclimbing procedure by a crossover operation between different members of a population of candidate solutions. By recombining the different color groups already seen before by the learning algorithm we expect  the approximation of $\mathbb{E}[f_A(S)]$, given by $f_{\theta}(S)$, to be more precise. 
\end{enumerate}

The pseudo-code of the proposed new deep learning guided memetic framework for graph coloring  (DLMCOL) is shown in Algorithm \ref{algo_memetic}.

\begin{algorithm}
\caption{Deep learning guided memetic framework }\label{algo_memetic}
\begin{algorithmic}[1]
%\STATE \
\STATE \bf{Input}: \normalfont{Graph $G = (V, E)$, population size $p$.}
\STATE \bf{Output}: \normalfont{The best legal  coloring $S^*$ found so far}
\STATE \normalfont{$P = \{S_1,\dots,S_p\}$ $\leftarrow$ population\_initialization}
\hfill $/*$ Section \ref{sec:InitGreedy}
\STATE Initialize the neural network $f_{\theta}$ with random weights.
\STATE $S^*= \emptyset$ and $f(S^*)= \infty$
\STATE $\{S^O_1,\dots,S^O_p\} \leftarrow \{S_1,\dots,S_p\}$
\REPEAT
\FOR {$i = \{1, \dots, p \}$, \textbf{in parallel}}
\STATE $S'_i \leftarrow \text{local\_search}(S^O_i)$ \hfill $/*$ Section \ref{sec:IteratedTabu}
\ENDFOR
\STATE $S'^* = \text{argmin} \{f(S'_i), i= 1, \dots, p \}$
\IF{$f(S'^*) < f(S^*)$}
\STATE $S^* \leftarrow S'^*$
\ENDIF
\STATE Build supervised learning training dataset $\mathcal{D} =\{(S^O_i,f(S'_i))\}_{i=1}^p\}$ and train the neural network $f_{\theta}$ on it.  \hfill $/*$ Section \ref{sec:DeepLearning}.
\STATE $D \leftarrow \text{distance\_computation} (S_1, \dots, S_p,S'_1, \dots, S'_p )$ \hfill $/*$ Section \ref{DistanceComputation}
\STATE $\{S_1, \dots, S_p \} \leftarrow \text{pop\_update} (S_1, \dots, S_p,S'_1, \dots, S'_p , D)$ \hfill $/*$ Section \ref{PopulationUpdate}
\STATE $\{S^O_1, \dots, S^O_p \} \leftarrow \text{build\_and\_select\_offsprings} (S_1, \dots, S_p, f_{\theta}, D)$ \hfill $/*$ Section \ref{Crossovers}
\UNTIL{stopping condition met}
\RETURN $S^*$
\end{algorithmic}
\end{algorithm}

The  algorithm  takes  a   graph $G$ as  input  and  tries  to  find  a  legal coloring $S$ with the minimum score $f(S)$. At the beginning, all the individuals of the population are initialized in parallel using a greedy random algorithm (cf. Section \ref{sec:InitGreedy}) and the neural network $f_{\theta}$ is initialized with random weights. Then, the algorithm repeats a loop (generation) until a stopping criterion (e.g., a cutoff time limit or a maximum number of generations) is met. Each generation $t$ involves the execution of five components:

\begin{enumerate}
\item The $p$ offspring individuals of the current population are simultaneously improved by running  in parallel $p$ local searches  on the GPU to find new legal solutions with a minimum score $f$ (cf. Section \ref{sec:IteratedTabu}).   For each of the $p$ improved individuals from step 1, we record  $S'_i$, the legal state with the lowest score $f(S'_i)$ obtained during each local search trajectory.
\item From these $p$ local search trajectories, a supervised learning training dataset $\mathcal{D} =\{(X_i,y_i)\}_{i=1}^p$ is built with $X_i = S^O_i$  and $y_i=f(S'_i)$ for $1 \leq i \leq p$ and the neural network $f_{\theta}$ is trained on this dataset during $N$ epochs (cf. Section \ref{sec:DeepLearning}).
\item The distances between all pairs of the existing individuals $\{S_1,\dots,S_p\}$ and new individuals $\{S'_1, \dots, S'_p\}$ are computed in parallel (cf. Section \ref{DistanceComputation}).
\item Then the population updating procedure (cf. Section \ref{PopulationUpdate}) merges the $2p$ existing and new individuals to create a new population of $p$ individuals, by taking into account the fitness $f$ of each individual and the distances between individuals in order to maintain some diversity in the population.
\item Finally each individual is matched with its $K$ nearest neighbors in the population. For each individual, $K$ offspring solutions are generated and the one with the best expected score evaluated with the neural network $f_{\theta}$ is selected (cf. Section \ref{Crossovers}). After this selection procedure, $p$ offspring individuals are selected and become the  $p$ new starting points $\{S^O_1,\dots,S^O_p\}$ which are improved in parallel by the local search procedure during the next generation ($t+1$).
\end{enumerate}

The algorithm stops when a predefined condition is reached and returns the best recorded solution $S^*$. The subsequent subsections present the five components of this deep learning guided memetic framework applied to the WVCP. In order to show some generality of the proposed framework, an application of this approach to the vertex coloring problem is presented in Section \ref{sec:Adaptation_GCP}.

\subsection{Initialization with a greedy random algorithm for the WVCP \label{sec:InitGreedy}}

In order to initialize the individuals of the population, we use a randomized greedy procedure which is known to be  very effective for the WVCP \cite{nogueira2021iterated,sun2018adaptive}.

First all the vertices are sorted in descending order of the weights and then in descending order of the degrees. Then a color is  assigned to each vertex without creating conflicts by randomly choosing  a color in the set of already used color. If no color is available for the vertex $i$ with weight $w_i$, a new color is created (and the score of the current solution is increased by $w_i$).

Notice that for the WVCP, the number of used colors to find a legal coloring $S$ minimizing the global score $f(S)$ is unknown in advance. However, it is at least strictly greater than the chromatic number of the graph $G$. In our case, we use a predefined maximum number of colors $k$ in order to specify the size  of the layers of the neural network and to allocate memory for the local searches on the GPU. Specifically, we set $k$ to be the maximum number of colors used in the initial solutions generated by the randomized greedy procedure. The new search space $\mathcal{S}$ restricted with the $k$ available colors is composed of the partitions of vertices $V$ into $k$ color groups:

\begin{equation*}
\mathcal{S}_k = \{ \{V_1,V_2,\dots,V_{k}\}: \cup_{i=1}^{k} V_i = V, V_i \cap V_j = \emptyset,i \neq j, 1 \leq i,j \leq k\}.
\end{equation*}

Note that the best solution found for each benchmark graph of the WVCP presented in Section \ref{sec:experiments} typically requires significantly less colors than $k$.

\subsection{Parallel iterated feasible and infeasible Tabu Search} \label{sec:IteratedTabu}

For local optimization, we employ a parallel iterated tabu search algorithm to simultaneously improve the individuals  of the current population. It relies on the adaptive feasible and infeasible tabu search procedure (AFISA) proposed in \cite{sun2018adaptive}, with some slight modifications. AFISA is a sequential procedure that improves a starting legal or illegal coloring by optimizing the fitness function $g$ given by:

\begin{equation}
g(S) = f(s) + \phi \times c(S)
\label{AFISA_score}
\end{equation}

\noindent where $\phi \in \mathbb{R}$ is an adaptive coefficient for the penalty function $c(S) = \sum_{\{u,v\} \in E} \delta_{uv}$ with:

\begin{equation}
    \delta_{uv} = \begin{cases}
      1 & \text{if $u \in V_i$, $v \in V_j$ and $i=j$ and $i\neq 0$}\\
      0 & \text{otherwise}
    \end{cases} 
\end{equation}

 AFISA improves the current coloring by successively changing the color of a vertex in the search space $\mathcal{S}_k$ (with a maximum of $k$ colors).  Such a change is called an one-move. To prevent the search from revisiting already visited colorings, a vertex cannot change its color for the next $tt$ (called tabu tenure) iterations\footnote{In the original AFISA algorithm, the tabu tenure concerns past moves (as in the original TabuCol algorithm \cite{hertz1987using}) instead of completely freezing a vertex. However we empirically observed that it is more effective to freeze a vertex that has just changed its color in order to avoid too much color changes of the same vertex without any improvement of the score (plateau).}. The tabu tenure is set to be $L + 0.2 \times |V|$, where $L$ is a random integer from $[0;9]$ and $|V|$ is the number of vertices in the graph.

Like in the AFISA algorithm, we perform successive searches by changing dynamically the value of the parameter $\phi$ in order to navigate in the space of legal and illegal colorings. The maximum number of successive local searches is set to $maxLSIters=10$. At the beginning the parameter is set to the value $k/(2|V|)* \max_{i=1,...,|V|} w_i$) for each individual $j$ of the population. At the end of each successive tabu search, if the best current solution $S_j$ found by the tabu search procedure is legal ($c(S_j)=0$), then $\phi$ is divided by 2 (in order to  increase  the  chance  of  visiting infeasible solutions); otherwise  $\phi$ is multiplied by $2$ (in order to  guide  the  search  toward  feasible regions)\footnote{In the original AFISA algorithm, $\phi$ is initially set to 1 and cannot be lower than 1. The adaptive mechanism only increases or decreases its value by 1.
We empirically observed that dividing or multiplying its value by 2 (instead of simply changing its value by 1) appears to be more effective for faster adjustments, especially for graphs with heavy weights.}.  For the last iteration of this iterative tabu search algorithm, we set $\phi = 2 \times \max_{i=1,...,|V|} w_i$) to make sure that each tabu search is forced at least one time toward a legal solution.

The pseudo code of the parallel iterative tabu search is shown in Algorithm \ref{algo_ITS}, which runs on the GPU to raise the quality of the current population in parallel. All the data structures required during the search are stored in each local thread memory running tabu search except the information of the graph which is stored in the global memory and shared by all tabu search runs.

\begin{algorithm}[h]
\caption{Parallel iterated tabu search with feasible and infeasible solutions}\label{algo_ITS}
\begin{algorithmic}[1]
\STATE \bf{Input}: \normalfont{Population  $P= \{S^O_1, \dots, S^O_p \}$, depth of tabu search $nbIter_{TS}$, maximum number of successive local searches $maxLSIters$, weights $w_i$ of each vertex $i$. }
\STATE \bf{Output}: \normalfont{Improved population $P'= \{S'_1, \dots, S'_p \}$}.
%\STATE  \textbf{for} $i = \{1, \dots, p \}$ \textbf{do, in parallel}
\FOR {$i = \{1, \dots, p \}$, \textbf{in parallel}}
\STATE  $S'_i= \emptyset$ and $f(S'_i)= \infty$ \hfill $/*$ Records the best solution found so far on each local thread.
\ENDFOR
%\STATE \textbf{end for}
\STATE $iter = 0$
\WHILE{$iter < maxLSIters$}
\FOR {$i = \{1, \dots, p \}$, \textbf{in parallel}}
\STATE $S_i \leftarrow \text{feasible\_and\_infeasible\_tabu\_search}(S_i,nbIter_{TS},\phi_i)$ \\
$/*$ Improve the solutions by running the tabu search procedure to minimize the extended fitness function $g$ with search depth  $nbIter_{TS}$ and penalization coefficient $\phi_i$.\\
\IF{$f(S_i) < f(S'_i)$ and $c(S_i)=0$}
\STATE $S'_i \leftarrow S_i$
\ENDIF
\IF{$iter < maxLSIters - 1$}
\IF{$c(S_i)=0$}
\STATE $\phi_i = \phi_i/2$
\ELSE
\STATE $\phi_i = \phi_i \times 2$
\ENDIF
\ELSE
\STATE $\phi_i = 2 \times \underset{i\in 1,\dots |V|}{max} w_i$
\ENDIF
\ENDFOR
\STATE $iter = iter + 1$
\ENDWHILE
\STATE \textbf{return} $P'= \{S'_1, \dots, S'_p \}$
\end{algorithmic}
\end{algorithm}

\subsection{Deep neural network training} \label{sec:DeepLearning}

Once all the parallel tabu searches are done, we collect the starting states $S^O_i$ and the best score $f(S'_i)$ found on each local thread. These data are then used to build a supervised training dataset $\mathcal{D} = \{S^O_i,f(S'_i)\}_{i=1}^p$ with $p$ examples whose entries are the $S^O_i$ in $\mathcal{S}_k$ and the corresponding targets are real values $f(S'_i)$.

A neural network $f_{\theta}: \mathcal{S}_k \rightarrow \mathbb{R}$, parametrized by a vector $\theta$ (initialized at random at the beginning), is successively trained on each new dataset $\mathcal{D}$ produced at each generation (online training) in order to be able to be more and more accurate at predicting the  expected score obtained after the local search procedure for any new starting point $S \in \mathcal{S}_k$. 

This neural network $f_{\theta}$ takes directly as input a coloring $S$ as a set of $k$ vectors $V_j$, $S=\{V_1, \dots, V_{k}\}$, where each $V_j$ is a binary vector of size $n$ indicating if the vertex $i$ belongs to the color group $j$. For such an entry $S$, the neural network outputs a real value  noted $f_\theta(S) \in \mathbb{R}$.

For the WVCP, one important characteristic of our neural network $f_{\theta}$ is that it is invariant by permutation of the group of colors of any solution $S$ given as input. It should be a function $f_{\theta}$ from $\mathcal{S}_k$ to $\mathbb{R}$ such that for any permutation $\sigma$ of the input color groups:

\begin{equation}
 f_{\theta}(V_{\sigma(1)}, \dots, V_{\sigma(k)}) = f_{\theta}(V_1, \dots, V_{k}).
\end{equation}

As indicated in \cite{lucas2018mixed,NIPS2017_f22e4747}, such permutation invariant functions can be obtained by combining the  treatments of each color group vector $V_j$ with an additional "color-averaging" operation that performs an average of the features across the different color groups. It has notably been shown in \cite{lucas2018mixed} that such operations are sufficient to recover all invariant functions from $\mathcal{S}_k$ to $\mathbb{R}$.

Using the notations proposed in \cite{lucas2018mixed}, for a coloring $S=\{V_1, \dots, V_{k}\}$, the color group invariant network $f_\theta$ is defined as:

\begin{equation}
    f_\theta(S) = \frac{1}{k} \sum_{i=1}^{k} (\phi_{\theta_P} \circ \phi_{\theta_{P-1}} \circ \dots \circ \phi_{\theta_{0}} (S))_i
\end{equation}

\noindent where each $\phi_{\theta_j}$ is a permutation invariant function from $\mathbb{R}^{k \times l_{j-1}}$ to $\mathbb{R}^{k \times l_{j}}$, where $l_j$'s are the layer sizes. Note that for the first layer $l_{j-1} = |V|$.

Each layer operation $\phi_{\theta}$ with $l$ input features and $L$ output features includes a weight matrix $\Lambda \in \mathbb{R}^{l \times L}$ that treats each color group independently,  a color-mixing weight matrix  $\Gamma \in \mathbb{R}^{l \times L}$ and a bias vector $\beta \in \mathbb{R}^{l}$.

As in classical multi-layer feed-forward neural networks, $\Lambda$ processes each color group vector of the solution $S$ independently. Then, the weight matrix $\Gamma$ processes the average vector $\rho(S)$ computed across the different $k$ color groups for each feature given by:

\begin{align}
   \rho(S)  & = \rho(V_1, \dots, V_{k}) \\
    & = \frac{1}{k}\sum_{i=1}^{k} V_i
\end{align}

The output of the layer $\phi_{\theta}$ is a matrix in $\mathbb{R}^{{k} \times L}$, which is the concatenation of ${k}$ output vectors of size $L$:

\begin{equation}
    \phi_{\theta}(S) = (\phi_{\theta}(S)_1, \dots, \phi_{\theta}(S)_{k})
\end{equation}

\noindent where for $1 \leq i \leq {k}$, 

\begin{equation}
   \phi_{\theta}(S)_i = \text{LeakyReLU}_{0.2}(\beta +  V_i \Lambda +  \rho(S) \Gamma).
\end{equation}

\noindent $\text{LeakyReLU}_{0.2}$ is a  non linear activation function defined as: $$\text{LeakyReLU}_{0.2}(x) = \text{max}(0.2 \times x, x).$$

After each local search procedure the neural network $f_\theta$ is trained during $N$ epochs on the new dataset $\mathcal{D}$ using Adam optimizer \cite{kingma2014adam} with initial learning rate $lr=0.001$ and batches of size $b=100$ in order to minimize the mean square error loss (MSE) between the outputs and the targets. In order to speed up the training, prior to the non linearity we apply a batch normalization layer \cite{ioffe2015batch}, adapted to keep the invariant property of the network. For the layer $j$ of the network, the ouput of the invariant batch normalization layer is:

\begin{equation}
    y = \frac{x - \mathbb{E}[x]}{ \sqrt{\text{Var}[x] + \epsilon}} * \gamma + \mu
\end{equation}

\noindent where $x$ is the input of size $(b,k,l_j)$. The mean and standard-deviation of $x$ are vectors of size $l_j$ calculated over the mini-batches and all the $k$ colors (in order to keep the invariance property). $\gamma$ and $\mu$ are learnable parameter in $\mathbb{R}$. $\epsilon=10^{-5}$ is a value added to the denominator for numerical stability. As in the original  work of \cite{ioffe2015batch}, during training the  batch normalization layer keeps running estimates of its computed mean and variance, which are then used for normalization during evaluation. The running estimates are kept with a default momentum of 0.1.
 
Once learned, this neural network will be used to select new crossovers for the next generation (see Section \ref{Crossovers} below), but before performing crossovers, we must decide  if the new  legal  colorings obtained after the parallel tabu search procedure can be inserted  into  the  population. For this purpose,  a  distance-and-quality based pool update strategy is used to create a new population satisfying  a  minimum  spacing  among  the  individuals  to  ensure  population  diversity \cite{PorumbelHK11}.  Maintaining  this  minimum  spacing requires  the  computation  of  pairwise  distances  between  the solutions, which is presented in the next subsection.

\subsection{Distance computation}  \label{DistanceComputation}

Following \cite{goudet2021population,lu2010memetic,porumbel2010evolutionary}, for population updating, we use a $p \times p$ matrix to record all the distances between any two solutions of the population. This symmetric matrix is initialized with the $p \times (p-1)/2$ pairwise distances computed  for each pair of individuals in the initial population, and then updated each time a new individual is inserted in the population.

To merge the $p$ new solutions and the $p$ existing solutions, we need to evaluate (i) $p \times p$  distances between each individual in the population $P=\{S_1,\dots,S_p\}$ and each improved individual in $P' = \{S'_1,\dots,S'_p\}$ and (ii) $p \times (p-1)/2$ distances between all the pairs of individuals in $P'$. All the $p\times p + p\times(p-1)/2$ distance computations are independent from one another, and are performed in parallel on the GPU (one computation per thread).

Given two colorings $S_i$ and $S_j$, we use the set-theoretic partition distance $D(S_i,S_j)$ to measure the dissimilarity between $S_i$ and $S_j$, which corresponds to the minimum number of vertices that need to be displaced between color classes of $S_i$ to transform $S_i$ to $S_j$ \cite{porumbel2011efficient}. The exact partition distance between two solutions can be calculated with the Hungarian algorithm \cite{kuhn1955hungarian} in $O(|V| + n^3)$ time. However, given that we need to compute millions of distances at each generation with the large population, we instead adopt the efficient approximation algorithm presented in \cite{porumbel2011efficient}, which scales in $O(|V|)$.

\subsection{Population update} \label{PopulationUpdate}

According to \cite{porumbel2010evolutionary,PorumbelHK11}, the population update procedure aims to keep the best individuals, but also to ensure a minimum spacing distance between the $p$ individuals. The update procedure is sequential, as we need to compare one by one existing individuals in the population $P_t = \{S_1, \dots, S_p \}$ at generation $t$ and the tabu search improved offspring solutions  in the population $P' = \{S'_1, \dots, S'_p\}$.

We use the population update procedure proposed in \cite{goudet2021population}. This procedure greedily adds one by one the best individuals of $P^{all} =\{S_1, \dots, S_p \} \cup  \{S'_1, \dots, S'_p\}$ in the population of the next generation $P_{t+1}$ until $P_{t+1}$ reaches $p$ individuals, such that $D(S_i,S_j) > |V|/10$ ($|V|$ is the number of vertices), for any $S_i,S_j \in P_{t+1}$, $i \neq j$. Each $D(S_i,S_j)$ corresponds to the approximation of the set-theoretic partition distance which was precomputed in the last step of the algorithm. 

\subsection{Parent matching and selection of crossovers with the neural network} \label{Crossovers}

At each generation, each individual of the population is matched with its $K$ nearest neighbors in the population (in the sense of the distance evaluated in subsection \ref{DistanceComputation}). We do not consider performing crossovers between  individuals too far away in the search space as this may result in poor quality offsprings (cf. \cite{moalic2018variations}). 

For each individual $i$, $K$ offspring solutions $S^j_i$ ($1 \leq j \leq K$) are generated using the well-known GPX crossover \cite{galinier1999hybrid,moalic2018variations}, where the individual $i$ is taken as the first parent and its neighbor as the second parent (the GPX crossover is not symmetric). 

For each individual $i$, among these $K$ crossovers, we select the one with the best expected score evaluated with the neural network of Section \ref{sec:DeepLearning}:

\begin{equation}
    S^0_i = \underset{S^j_i, 1 \leq j \leq K}{\text{argmin}}f_{\theta}(S_i^j)
    \label{eq:select_cross}
\end{equation}

 After this selection procedure, $p$ offspring solutions are identified that serve as the $p$ new starting points $\{S^O_1,\dots,S^O_p\}$ of the parallel tabu search search procedure during the next generation ($t+1$).

\subsection{Adaptations of the algorithm for the vertex coloring problem \label{sec:Adaptation_GCP}}

The vertex coloring problem COL aims at finding the smallest $k$ for a given graph $G$ (its chromatic number $\chi(G)$) such that $G$ admits a legal coloring using $k$ colors. Following the literature on graph coloring \cite{galinier2013recent}, we tackle this problem by solving a series of $k$-coloring problems ($k$-COL) with decreasing $k$ values. Starting from an initial number of $k$ colors, as soon as a legal $k$-coloring is found, $k$ is decreased by one. This process is repeated until no legal solution with $k$ colors can be found and the last $k$ admitting a legal $k$-coloring defines an upper bound of the chromatic number of the graph.  Let $k$ be the given colors and $S=\{V_1,V_2,\dots,V_k\}$ be a candidate $k$-coloring, the $k$-COL problem can be seen as the optimization problem that aims to minimize the number of conflicts given by $f(S)$ (until it reaches 0):

\begin{equation} \label{f}
    f(S) = \sum_{\{u,v\} \in E} \delta_{uv}
\end{equation}

\noindent where 

\begin{equation}
    \delta_{uv} = \begin{cases}
      1 & \text{if $u \in V_i$, $v \in V_j$ and $i=j$}\\
      0 & \text{otherwise}.
    \end{cases}  
\end{equation}

For this minimization problem, we use the same deep learning guided memetic framework presented in the last subsections. The only parts of the general framework that require specific adaptations for the $k$-COL concern the initialization and the parallel local search. For the $k$-COL, we use a pure random initialization procedure. At the beginning, for each individual of the population, each node $v \in V$ receives a random color in $\{1, \dots, k \}$. For the local search, we run in parallel the popular TabuCOL algorithm \cite{hertz1987using} (i.e., its efficient implementation presented in \cite{galinier1999hybrid}) on the GPU to raise the quality of the current population during $128 \times |V|$ iterations, where $|V|$ is the order of the graph. TabuCOL is launched with its default parameters setting like in \cite{moalic2018variations}.

\section{Experimental results \label{sec:experiments}}

This section is dedicated to a computational assessment of the proposed deep learning memetic framework  for solving the weighted vertex coloring  problem and the conventional vetex coloring problem, by making comparisons with state-of-the-art methods.

\subsection{Benchmark instances} \label{Benchmark}

We carried out extensive experiments on the WVCP benchmark graphs used in the recent studies \cite{nogueira2021iterated,sun2017feasible,wang2020reduction}: the pxx, rxx, DIMACS/COLOR small, and DIMACS/COLOR large instances. The pxx and rxx instances are based on matrix-decomposition problems \cite{prais2000reactive}, while DIMACS/COLOR small \cite{cornaz2017solving,furini2012exact} and DIMACS/COLOR large \cite{sun2017feasible} are based on DIMACS and COLOR competitions.

 As indicated in \cite{nogueira2021iterated,wang2020reduction}, for the WVCP, a preprocessing procedure can be applied to reduce a graph $G$ with the set of weight $W$. For each clique $C_l$ with $l$ vertices, if we note $w'$ the smallest weight of this set, all the vertices $i$ in the graph with a degree equal to $l-1$ and a weight $w_i < w'$ can be removed from the graph without changing the optimal WVCP score of this instance. Enumerating all the cliques of the graph is a challenging problem.  We used the \textit{igraph python package}\footnote{\url{https://igraph.org/python/}} with a timeout of ten seconds for all instances. For small instances it is enough to enumerate all the cliques of a graph. For our experimental evaluation, DLMCOL as well as all the competitors take these reduced graph as input. 

For the vertex coloring problem COL, we conducted experiments on the classical DIMACS benchmark graphs used in most of the best coloring methods for this problem \cite{moalic2018variations,titiloye2011quantum}. These instances can be separated into two categories: \textit{easy} instances and \textit{difficult} instances. For the \textit{easy} instances, most recent heuristics can reach the chromatic number (or its best known upper bound) in a short amount of time, while for the \textit{difficult} instances, no single algorithm is able to reach the chromatic number or the best known result for all these graphs. In this section, we only report the results for the most \textit{difficult} instances. The results for the \textit{easy} DIMACS instances are summarized in Appendix \ref{app:easyGCP}. 

For the WVCP and the COL, due to local memory limit on each thread of the GPU for the local searches, the DLMCOL algorithm was not runned on the biggest instances of the DIMACS/COLOR benchmarks (when $n \times k > 200  000$): C2000.5, C2000.9, DSJC1000.9, r1000.5 and wap01-4a.

\subsection{Implementation and parameter setting \label{sec:Parameter}}

The DLMCOL algorithm was coded in Python with the Numba 0.53.1 library for CUDA kernel implementation (local searches, distance computations, crossovers). The neural network was implemented in Pytorch  1.8.1. DLMCOL is specifically designed to run on GPUs.  In this work we used a V100 Nvidia graphic card with 32 GB memory. The code of DLMCOL is publicly available at  \url{https://github.com/GoudetOlivier/DLMCOL_WVCP}.

The population size $p$ of DLMCOL is set to $p=20480$, which is chosen as a multiple of the number of 64 threads per block. This large population size offers a good performance ratio on the Nvidia V100 graphics cards, while remaining reasonable for pairwise distance calculations in the population, as well as the memory occupation on the GPU for medium instances ($n \leq 500$).  However for large instances ($n > 500$ and $k > 90$), we set $p=8192$ in order to limit the global memory occupation on the device. 

The number of tabu iterations $nbIter_{TS}$ depends on the size $|V|$ of the graph.  The maximum number of iterated tabu searches launched at each generation, $LSIters$, is set to 10. The minimum spacing distance $MS$  used for pool update is set to $\frac{|V|}{10}$.
 
For the neural network we implemented an architecture  with 4 hidden layers of size $5|V|$, $2|V|$, $|V|$ and $|V|//2$ for the WVCP and 9 hidden layers of size $10|V|$, $5|V|$, $2|V|$, $2|V|$, $2|V|$, $2|V|$, $2|V|$, $|V|$ and $|V|//2$ for the $k$-COL problem.   The neural network is trained at each generation with Adam optimizer \cite{kingma2014adam} and initial learning rate $l_r=0.001$.
 
Tables \ref{table:parameters_DLM_WVCP} and \ref{table:parameters_DLM_COL}  summarize the parameter settings for the WVCP and the $k$-COL problems, which can be considered as the default and were used for all our experiments.

\begin{table}[!h]
\centering
\caption{Parameter setting in DLMCOL for the WVCP and the COL}
\begin{scriptsize}
\begin{tabular}{lll}
Parameter & Description & Value\\
\hline
$p$ & Population size & 20480 (8192)\\
$maxLSIters$ & Maximum number of successive local searches & 10\\
$nbIter_{TS}$ & Depth of tabu search & $10 \times |V|$\\
$\alpha$ & Tabu tenure parameter  & 0.2\\
$MS$ & Minimum  spacing between two individuals & $\frac{|V|}{10}$\\
$l_r$ & Learning rate of the neural network & 0.001
\\ $N$ & Number of epochs of the training & 20\\ 
$K$ & Number of considered neighbors  for crossover selection & 32 \\
\hline
\end{tabular}
\end{scriptsize}
\label{table:parameters_DLM_WVCP}
\end{table}

\begin{table}[!h]
\centering
\caption{Parameter setting in DLMCOL for the $k$-COL problem}
\begin{scriptsize}
\begin{tabular}{lll}
Parameter & Description & Value\\
\hline
$p$ & Population size & 20480 (8192)\\
$nbIter_{TS}$ & Depth of tabu search & $128 \times |V|$\\
$\alpha$ & Tabu tenure parameter  & 0.6\\
$MS$ & Minimum  spacing between two individuals & $\frac{|V|}{10}$\\
$l_r$ & Learning rate of the neural network & 0.001\\ 
$N$ & Number of epochs of the training & 5\\ 
$K$ & Number of considered neighbors  for crossover selection & 16 \\
\hline
\end{tabular}
\end{scriptsize}
\label{table:parameters_DLM_COL}
\end{table}

\subsection{Comparative results on weighted vertex coloring benchmarks \label{sec:benchmarks}}

This section shows a comparative analysis on the pxx, rxx, DIMACS/COLOR small, and DIMACS/COLOR large instances with respect to the state-of-the-art methods \cite{nogueira2021iterated,sun2017feasible,wang2020reduction}. The reference methods include the three best recent heuristics:  AFISA  \cite{sun2017feasible},  RedLS \cite{wang2020reduction} and ILS-TS \cite{nogueira2021iterated}. When they are available, we also include the optimal  scores obtained with the exact algorithm MWSS \cite{cornaz2017solving} and extracted from \cite{nogueira2021iterated}. 

Given the stochastic nature of the DLMCOL  algorithm, each instance was independently solved 10 times. For small instances presented in Tables \ref{results1}--\ref{results2}, a time limit of 1 hour was used. However for medium and large instance in Tables \ref{results3} and \ref{results4}, as training the neural network and performing all the tabu searches with the large population is time consuming, a cutoff limit of 48 hours was retained.

For a fair comparison, we also launched the reference methods RedLS \cite{wang2020reduction} and ILS-TS \cite{nogueira2021iterated} during 48 hours on a computer with an Intel Xeon E5-2670 processor (2.5 GHz and 2 GB RAM), until no improvement was observed. As the available AFISA binary code does not allow setting a cutoff time, we only report its results mentioned in the original article \cite{sun2017feasible}. However, we acknowledge that the comparison remains difficult in terms of computational time between DLMCOL and the competitors, as DLMCOL was run on GPUs while the other algorithms, AFISA, RedLS and ILS-TS used CPUs.  Therefore the timing information is provided for indicative purposes only.

Columns 1, 2, and 3  of Tables \ref{results1}--\ref{results4} show the characteristics of each instance (i.e., name of the instance, number of vertices $|V|$, and optimal score reported in the literature if available).  Columns 4-9 present the best and average scores  obtained by the reference algorithms, as well as the average time in second required to obtain their best results.  The results of the proposed DLMCOL algorithm are reported in columns 10 and 11. Boldfaced numbers show the dominating values while a star indicates a new upper bound\footnote{The certificates of the new best solutions from DLMCOL for the WVCP are available at \url{https://github.com/GoudetOlivier/DLMCOL_WVCP}.}.

The optima for the instances of Tables \ref{results1}-\ref{results3} are known except for four instances DSJC125 and two instances R100. As the result, no algorithm can further improve these bounds. For these graphs, the proposed algorithm and the latest ILS-TS algorithm report the same results and both algorithms perform better than AFISA and RedLS. However, the computation time required by DLMCOL to achieve its results is in general higher that the reference algorithms in particular when compared with ILT-TS. This is not really surprising given that the neural network training requires additional computation time in addition of the time needed by the optimization components.

For the larger instances reported in Table \ref{results4}, DLMCOL obtains excellent results by reaching the best-known score for 31 over 49 instances. For 11 of them, DLMCOL even finds new upper bounds that were never been reported before. In particular, improvements are quite important for 3 instances with a high reduction of the best-known scores: DSJC500.5 from 707 to 685, flat1000\_50\_0 from 1184 to 924 and latin\_square\_10 from 1542 to 1480.

However, DLMCOL does not work well for large sparse graphs with low edge density such as DSJC1000.1, inithhx.i.2, inithhx.i.3 and wapXXa. For these graphs, it seems that it is very hard for the neural network to learn a common backbone of good solutions. An analysis of these negative results is proposed in Section \ref{sec:analysis}.

For the largest graphs, we notice that the DLMCOL algorithm converges slowly, but continually. Even after 48 hours, DLMCOL still improves its solutions. This indicates that the algorithm is not trapped in local optima, which is a common problem for most existing WVCP algorithms. For the graphs DJSC1000.5, flat1000\_60\_0 and flat1000\_76\_0,  DLMCOL was able to obtain still better new upper bounds of 1185, 1162, 1165, after 138, 98 and 95 hours, respectively, raising the total number of improved upper bounds to 14 for the WVCP.

For the large instances in Table \ref{results4}, DLMCOL and the best competitors RedLS and ILS-TS have their own advantage respectively, while the proposed algorithm has the best overall success rate of 63\% against AFISA (10\%), RedLS (53\%) and ILS-TS (49\%).

\begin{table*}[!h]
\centering
\tiny
\caption{Comparative results of DLMCOL  with the state-of-the-art methods (AFISA, RedLS, ILS-TS) for DIMACS/COLOR small instances of the WVCP. Dominating  results are indicated in boldface.   \label{results1}}
\begin{tabular}{lll|ll|ll|ll|ll}
   \hline
 \multicolumn{3}{c|}{Instance}   & \multicolumn{2}{|c|}{AFISA} & \multicolumn{2}{|c|}{RedLS} & \multicolumn{2}{|c|}{ILS-TS} & \multicolumn{2}{|c}{DLMCOL} \\
   \hline
  Graph name & $|V|$  & Opti.  & Best (Avg.) & t (s) & Best (Avg.) & t (s) & Best (Avg.) & t (s)  &  Best (Avg.) & t (s)   \\
    \hline
DSJC125.1g & 124 & -  & \textbf{23} (24)  & 3016 & \textbf{23}  & 0.01 & \textbf{23}  & 3 & \textbf{23}  & 27  \\
DSJC125.1gb & 124  & -  & \textbf{90} (92.5) &  402 & 91 (91.7) & 696 & \textbf{90} & 15 & \textbf{90}  & 28  \\ 
DSJC125.5g & 125  & -    & \textbf{71} (72.3) &  216 & 72  & 32895 & \textbf{71}  & 77 & \textbf{71}  & 183  \\ 
DSJC125.5gb & 125  & -   & 243 (250.2) &  369 & 241 (241.3) & 13528 & \textbf{240}  & 219 & \textbf{240}  & 40  \\ 
DSJC125.9g & 125  & 169  & \textbf{169} (169.9) &  16 & \textbf{169} & 3493 & \textbf{169}  & 1.27 & \textbf{169}  & 51  \\
DSJC125.9gb & 125  & 604  & \textbf{604} (605.5) &  444 & \textbf{604} & 17.62 & \textbf{604}  & 59.8 & \textbf{604}  & 51  \\
GEOM100 & 100  & 65  & \textbf{65} (65.0) &  0.81 & \textbf{65} (67.5) & 0.01 & \textbf{65}  & 0.03 & \textbf{65} & 27  \\
GEOM100a & 100  & 89  & \textbf{89} (89.5)  &  110 & 90 (93.3) & 0.01 & \textbf{89}  & 0.65 & \textbf{89} & 30  \\
GEOM100b & 100  & 32  & \textbf{32} (33.1)  &  59 & \textbf{32} & 0.6 & \textbf{32} & 0.02 & \textbf{32}  & 18  \\
GEOM110 & 110  & 68  & \textbf{68} (68.0)  & 34 & \textbf{68} (69.9) & 0.03 & \textbf{68}  & 0.05 & \textbf{68}  & 19  \\
GEOM110a & 110  & 97  & \textbf{97} (97.8) & 177 & \textbf{97} (99.4)  & 0.04 & \textbf{97}  & 0.58 & \textbf{97}  & 20  \\
GEOM110b & 110  & 37 & \textbf{37} (37.9)  &  131 & \textbf{37} (37.71) & 22.1 & \textbf{37}  & 0.25 & \textbf{37}  & 27  \\
GEOM120 & 120  & 72  & \textbf{72}  & 33 & \textbf{72} (73.1) & 9.0 & \textbf{72}  & 0.03 & \textbf{72}  & 25  \\
GEOM120a & 120  & 105  & \textbf{105} (106.3) & 156 & \textbf{105} (105.9) & 1.96 & \textbf{105}  & 0.77 & \textbf{105}  & 30  \\
GEOM120b & 120  & 35  & \textbf{35} (37.3) & 67.7 & \textbf{35} (35.25) & 14.38 & \textbf{35}  & 0.81 & \textbf{35}  & 33  \\
GEOM30b & 30  & 12   & \textbf{12}  &  0.02 & \textbf{12} & 0.01 & \textbf{12}  & 0.01 & \textbf{12} & 20  \\
GEOM40b & 40   & 16 & \textbf{16}  & 0.03 & \textbf{16} (16.6) & 0.01 & \textbf{16}  & 0.01 & \textbf{16} & 20  \\
GEOM50b & 50  & 18 & \textbf{18}  & 0.02 & \textbf{18} (18.2) & 62.02 & \textbf{18}  & 0.01 & \textbf{18} & 15 \\
GEOM60b & 60  & 23 & \textbf{23}  & 0.22 & \textbf{23}  & 0.01 & \textbf{23}  & 0.01 & \textbf{23} & 20  \\
GEOM70 & 70  & 47  & \textbf{47}  &  5 & \textbf{47} (48.6) & 0.01 & \textbf{47}  & 0.02 & \textbf{47} & 19  \\
GEOM70a & 70  & 73 & \textbf{73} & 4 & \textbf{73} (73.6) & 0.25 & \textbf{73}  & 0.03 & \textbf{73} & 20 \\
GEOM70b & 70  & 24 & \textbf{24}  & 12 & \textbf{24}  & 15.1 & \textbf{24}  & 0.01 & \textbf{24}  & 24  \\
GEOM80 & 80  & 66 & \textbf{66}  & 2 & 67 (67.4) & 0.01 & \textbf{66}   & 0.01 & \textbf{66}  & 20  \\
GEOM80a & 80  & 76 & \textbf{76} (76.1)  & 137 & \textbf{76} (78.2) & 1.3 & \textbf{76}   & 0.04 & \textbf{76}  & 23  \\
GEOM80b & 80  & 27 & \textbf{27}  (27.8) &  67 & \textbf{27}  & 24
 & \textbf{27}   & 0.06 & \textbf{27}  & 12  \\
GEOM90 & 90  & 61 & \textbf{61} (61.2)  & 89 & \textbf{61} (63.5) & 1.74 & \textbf{61}  & 0.15 & \textbf{61}  & 21  \\
GEOM90a & 90  & 73 & \textbf{73} (74) &  512 & \textbf{73} (74.1) & 3.65 & \textbf{73}   & 0.55 & \textbf{73}  & 23  \\
GEOM90b & 90  & 30 & \textbf{30} (30.1)  &  67 & \textbf{30} (30.1) & 0.17 & \textbf{30}  & 0.02 & \textbf{30}  & 25  \\
R100\_1g & 100  & 21 & \textbf{21} (22)  & 114 & \textbf{21} (21.8) & 508.0 & \textbf{21}  & 7.14 & \textbf{21} & 300  \\
R100\_1gb & 100  & 81 & \textbf{81} (83.8) &  3 & \textbf{81} (81.4) & 1279.7 & \textbf{81}  & 1.98 & \textbf{81}  & 27  \\
R100\_5g & 100   & - & \textbf{59} (60.1) &  7 & \textbf{59} & 193 &  \textbf{59}  & 0.2 & \textbf{59}  & 28  \\
R100\_5gb & 100  & - & 221 (224.1) & 187 & \textbf{220} (222) & 687 & \textbf{220}  & 4 & \textbf{220} & 26  \\
R100\_9g & 100   & 141 & \textbf{141} (141.3)  & 21 & \textbf{141}  & 0.62 & \textbf{141}  & 40.8 & \textbf{141}  & 36  \\
R100\_9gb & 100  & 518 & \textbf{518} (5449.3) & 1152 & \textbf{518}  & 8.17 & \textbf{518} (518.3) & 1066.1 & \textbf{518}  & 35  \\
R50\_1g & 50  & 14  & \textbf{14}  &  0.14  & \textbf{14} (14.1) & 0.01 & \textbf{14}  & 0.01 & \textbf{14} &  19 \\
R50\_1gb & 50  & 53 & \textbf{53} (53.0) &  0.24 & \textbf{53} (53.1) & 0.07 & \textbf{53}   & 0.01 & \textbf{53}  & 19  \\
R50\_5g & 50  & 37 & \textbf{37} (37.0)  &  0.95 & \textbf{37}  & 0.01 & \textbf{37}  & 0.02 & \textbf{37}  & 24  \\
R50\_5gb & 50  & 135 & \textbf{135} (135.3) & 4 & \textbf{135}  & 0.09 & \textbf{135}  & 0.21 & \textbf{135}  & 20  \\
R50\_9g & 50  & 74 & \textbf{74}  & 1 & \textbf{74}  & 0.02 & \textbf{74}  & 0.01 & \textbf{74} & 21  \\
R50\_9gb & 50  & 262 & \textbf{262} & 13 & \textbf{262}  & 0.01 & \textbf{262} & 1.4 & \textbf{262}  & 22  \\
R75\_1g & 75  & 18 & \textbf{18} (18.4) & 11 & \textbf{18}  & 2.62 & \textbf{18} & 0.28 & \textbf{18} & 20  \\
R75\_1gb & 75  & 70 & \textbf{70} (70.1) &  2 & \textbf{70} (72.4) & 0..35 & \textbf{70}  & 0.23 & \textbf{70}  & 20  \\
R75\_5g & 75  & 51 & \textbf{51} (51.4) &  01 & \textbf{51} (51.2) & 598.6 & \textbf{51}  & 0.39 & \textbf{51}  & 22  \\
R75\_5gb & 75  & 186 & \textbf{186} (189) & 19 & \textbf{186} & 51.1 & \textbf{186}  & 2 & \textbf{186}  & 23  \\
R75\_9g & 75  & 110 & \textbf{110} & 3 & \textbf{110} & 0.08 & \textbf{110}  & 0.1 & \textbf{110} & 27  \\
R75\_9gb & 75  & 396 & \textbf{396} (396.4)  & 146 & \textbf{396}  & 0.42 & \textbf{396}  & 3.9 & \textbf{396} & 26  \\
\hline
 & &  &  & & & & & &  \\
 Best rate &  & & 95\% & & 89\% & & 100\%  & & 100\% \\
\hline
\end{tabular}
\end{table*}

\begin{table*}[!h]
\centering
\tiny
\caption{Comparative results of DLMCOL  with the state-of-the-art methods (AFISA, RedLS, ILS-TS) for pxx instances of the WVCP. Dominating  results are indicated in boldface.  \label{results2}}
\begin{tabular}{lll|ll|ll|ll|ll}
   \hline
 \multicolumn{3}{c|}{Instance}   & \multicolumn{2}{|c|}{AFISA} & \multicolumn{2}{|c|}{RedLS} & \multicolumn{2}{|c|}{ILS-TS} & \multicolumn{2}{|c}{DLMCOL} \\
   \hline
  Graph name & $|V|$  & Opti. & Best (Avg.) & t (s) & Best (Avg.) & t (s) & Best (Avg.) & t (s)  &  Best (Avg.) & t (s)   \\
    \hline
P06 & 16  & 565 & \textbf{565}  &  0.0 & \textbf{565}  & 0.01 & \textbf{565}  & 0.01 & \textbf{565}  & 21  \\
P07 & 24  & 3771 & \textbf{3771}  &  0.0 & \textbf{3771} (3773.5) & 0.01 & \textbf{3771} & 0.01 & \textbf{3771} & 18  \\
P08 & 24  & 4049 & \textbf{4049}   & 0.2 & \textbf{4049}  & 0.03 & \textbf{4049} & 0.21 & \textbf{4049}  & 18  \\
P09 & 25  & 3388 & \textbf{3388} (3388.2) &  1 & \textbf{3388}  & 0.01 & \textbf{3388} & 0.01 & \textbf{3388}  & 20  \\
P10 & 16  & 3983 & \textbf{3983}  &  0.7 & \textbf{3983}  & 0.01 & \textbf{3983} & 0.01 & \textbf{3983}  & 18  \\
P11 & 18  & 3380 & \textbf{3380} & 0.0 & \textbf{3380}  & 0.01 & \textbf{3380} & 0.01 & \textbf{3380}  & 18  \\
P12 & 26  & 657 & \textbf{657} & 0.0 & \textbf{657}  & 0.01 & \textbf{657} & 0.01 & \textbf{657}  & 19  \\
P13 & 26  & 3220 & \textbf{3220} (3221.1) & 0.7 & \textbf{3220} (3229) & 0.01 & \textbf{3220} & 0.09 & \textbf{3220}  & 20  \\
P14 & 31  & 3157 & \textbf{3157} & 0.0 & \textbf{3157}  & 0.01 & \textbf{3157} & 0.01 & \textbf{3157} & 19  \\
P15 & 34  & 341 & \textbf{341}  & 1.8 & \textbf{341} (343.1) & 0.01 & \textbf{341} & 0.01 & \textbf{341}  & 24  \\
P16 & 34  & 2343  & \textbf{2343} &  0.7 & \textbf{2343} (2383.3) & 0.01 & \textbf{2343} & 0.01 & \textbf{2343}  & 21  \\
P17 & 37   & 3281 & \textbf{3281} (3322.2) &  2.7 & \textbf{3281} (3282.6) & 0.05 & \textbf{3281} & 0.01 & \textbf{3281} & 21  \\
P18 & 35  & 3228 & \textbf{3228} & 0.1 & \textbf{3228}  & 0.01 & \textbf{3228} & 0.01 & \textbf{3228}  & 22  \\
P19 & 36  & 3710 & \textbf{3710} & 0.4 & \textbf{3710}  & 0.01 & \textbf{3710} & 0.01 & \textbf{3710}  & 22  \\
P20 & 37  & 1830 & \textbf{1830} (1841) & 4.9 & \textbf{1830} (1844) & 0.05 & \textbf{1830} & 0.38 & \textbf{1830}  & 22  \\
P21 & 38  & 3660 & \textbf{3660} (3660.5) &  0.8 & \textbf{3660} (3707.0) & 0.01 & \textbf{3660} & 0.01 & \textbf{3660}   & 23  \\
P22 & 38  & 1912 & \textbf{1912} (1912.2) & 0.3 & \textbf{1912} (1946) & 0.21 & \textbf{1912} & 0.01 & \textbf{1912}  & 19  \\
P23 & 44  & 3770 & \textbf{3770} (3793.0) & 0.3 & \textbf{3770} (3804) & 0.04 & \textbf{3770} & 0.01 & \textbf{3770}  & 19  \\
P24 & 34  & 661 & \textbf{661} & 0.0 & \textbf{661} (667.1) & 0.19 & \textbf{661} & 0.01 & \textbf{661}  & 18  \\
P25 & 36 & 504  & \textbf{504} & 0.3 & \textbf{504}  & 0.01 & \textbf{504} & 0.01 & \textbf{504} & 19  \\
P26 & 37 & 520  & \textbf{520} & 0.1 & \textbf{520}  & 0.01 &  \textbf{520} & 0.01 & \textbf{520}  & 19  \\
P27 & 44   & 216 & \textbf{216}  &  0.1 & \textbf{216} (219) & 0.01 & \textbf{216} & 0.07 & \textbf{216}  & 20 \\
P28 & 44   & 1729 & \textbf{1729} (1735.1) & 2.6 & \textbf{1729}  & 0.01 & \textbf{1729} & 0.01 & \textbf{1729}  & 19  \\
P29 & 53  & 3470 & \textbf{3470}  &  0.1 & \textbf{3470}  & 0.01 & \textbf{3470} & 0.01 & \textbf{3470}  & 19  \\
P30 & 60  & 4891 & \textbf{4891} &  54 & \textbf{4891} (4901) & 0.01 & \textbf{4891} & 0.01 & \textbf{4891}  & 20  \\
P31 & 47 & 620 & \textbf{620} & 3.7 & \textbf{620}  & 0.01 & \textbf{620} & 0.01 & \textbf{620}  & 18  \\
P32 & 51  & 2480 & \textbf{2480} &  0.4 & \textbf{2480}  & 0.01 & \textbf{2480} & 0.01 & \textbf{2480}   & 20  \\
P33 & 56  & 3018 & \textbf{3018} (3029.7) & 0.4 & \textbf{3018} (3096) & 0.02 & \textbf{3018} & 0.01 & \textbf{3018}  & 22  \\
P34 & 74  & 1980 & \textbf{1980} (1980.5) & 3.0 & \textbf{1980} (1994) & 0.01 & \textbf{1980} & 0.03 & \textbf{1980} & 26  \\
P35 & 86  & 2140  & \textbf{2140} (2145.0) &  4.5 & \textbf{2140} (2161) & 0.01 & \textbf{2140} & 0.02 & \textbf{2140}  & 23  \\
P36 & 101  & 7210  & \textbf{7210} (7385) & 0.1 & \textbf{7210} & 0.01 & \textbf{7210} & 0.01 & \textbf{7210} & 27  \\
P38 & 87  & 2130 & \textbf{2130} (2139.5) &  9.5 & 2140 (2161) & 0.01 & \textbf{2130} & 0.29 & \textbf{2130}  & 25  \\
P40 & 86  & 4984 & \textbf{4984}  (5016.6) & 5.1 & 5005 (5082.7) & 0.01 & \textbf{4984} & 0.33 & \textbf{4984}  & 25  \\
P41 & 116  & 2688 & \textbf{2688} ( 2688.1) & 0.1 & \textbf{2688} (2785.8) & 0.04 & \textbf{2688} & 0.32 & \textbf{2688}  & 453  \\
P42 & 138  & 2466 & \textbf{2466} (2671.2) &  931.0 & 2482 (2539.9) & 0.02 & \textbf{2466} & 9.02 & \textbf{2466}  & 35  \\
\hline
 & & & &  & & & & & &  \\
 Best rate & & & & 100\% & & 91\% & & 100 \% & & 100\% \\
\hline
\end{tabular}
\end{table*}

\begin{table*}[!h]
\centering
\tiny
\caption{Comparative results of DLMCOL  with the state-of-the-art methods (AFISA, RedLS, ILS-TS) for rxx instances of the WVCP. Dominating  results are indicated in boldface.  \label{results3}}
\begin{tabular}{lll|ll|ll|ll|ll}
   \hline
 \multicolumn{3}{c|}{Instance}   & \multicolumn{2}{|c|}{AFISA} & \multicolumn{2}{|c|}{RedLS} & \multicolumn{2}{|c|}{ILS-TS} & \multicolumn{2}{|c}{DLMCOL} \\
   \hline
  Graph name & $|V|$  & Opti. & Best (Avg.) & t (s) & Best (Avg.) & t (s) & Best (Avg.) & t (s)  &  Best (Avg.) & t (s)   \\
    \hline
r01 & 144  & 6724 & \textbf{6724} (6727.8) & 49.5  & 6732 (6769.2) & 0.01 & \textbf{6724} & 0.96 & \textbf{6724}  & 36  \\
r02 & 142  & 6771 & \textbf{6771} (6780.6)  & 85.3 & 6774 (6818.6) & 0.01 & \textbf{6771} & 0.25 & \textbf{6771}  & 35  \\
r03 & 139 & 6473 & \textbf{6473} (6490.8) & 190.2 & 6505 (6597.7) & 233.5 & \textbf{6473} & 4.53 & \textbf{6473} & 33  \\
r04 & 151  & 6342 & \textbf{6342} (6403.2) &  467.3 & 6349 (6427.7) & 0.42 & \textbf{6342} & 0.83 & \textbf{6342}  & 38  \\
r05 & 142  & 6408 &  \textbf{6408} (6466.3) &  71.7 & 6411 (65010.3) & 0.42 & \textbf{6408} & 2.57 & \textbf{6408} & 38  \\
r06 & 148  & 7550 & \textbf{7550} (7555.9) & 29.2 & \textbf{7550} (7558.9) & 0.01 & \textbf{7550} & 0.01 & \textbf{7550}  & 38 \\
r07 & 141  & 6889 & \textbf{6889} (7555.9) & 34.8  & 6910 (6974.2) & 954 & \textbf{6889} & 7.29 & \textbf{6889}  & 37  \\
r08 & 138  & 6057 & \textbf{6057} (6080.3) &  311.7 & 6071 (6147.4) & 0.04 & \textbf{6057} & 1.6 & \textbf{6057}  & 234  \\
r09 & 129   & 6358 & \textbf{6358} (6393.8) & 395.2 & 6390 (6451.9) & 66.13 & \textbf{6358} & 1.84 & \textbf{6358}  & 35  \\
r10 & 150  & 6508 & \textbf{6508} (6519.3) &  461.1 & 6518 (65078.6) & 0.05 & \textbf{6508} & 2.46 & \textbf{6508} & 39  \\
r11 & 208  & 7654  & \textbf{7654} (7710.6) &  9542.2 &  7691 (7739.5) & 489.45 & \textbf{7654} & 5.27 & \textbf{7654} & 432  \\
r12 & 199   & 7690 & 7691 (7710.4) &  9542.2 &  7694 (7730.2) & 2.61 & \textbf{7690} & 4.33 & \textbf{7690} & 58  \\
r13 & 217  & 7500 & 7521 (7558.3) &  619.5 &  7524 (7566.7) & 0.04 & \textbf{7500} & 5.8 & \textbf{7500} & 66  \\
r14 & 214  & 8254 & \textbf{8254} (8283.9) &  8044.1 & 8288 (8371.4) & 0.87 & \textbf{8254} & 4.78 & \textbf{8254}  & 60  \\
r15 & 198  & 8021 & \textbf{8021} (8126.8) &  2559.1 & \textbf{8021} (8024.0) & 0.01 & \textbf{8021} & 0.01 & \textbf{8021}  & 54  \\
r16 & 188  & 7755 & \textbf{7755} (7789.2) &  195.5 &  7764 (7809.4) & 0.01 & \textbf{7755} & 11.22 & \textbf{7755}  & 51  \\
r17 & 213  & 7979 & \textbf{7979} (8030.3) &  855.4 & 8011 (8064.3) & 0.86 & \textbf{7979} & 4.39 & \textbf{7979}  & 242  \\
r18 & 200  & 7232 & \textbf{7232} (7278.9) &  868.2 &  7240 (7295.3) & 11.53 & \textbf{7232} & 26.4 & \textbf{7232} & 4374  \\
r19 & 185  & 6826 & 6840 (6868.1) &  395.5 & \textbf{6826} (6850.5) & 39.15 & \textbf{6826} & 2.09 & \textbf{6826}  & 189  \\
r20 & 217  & 8023 & \textbf{8023} (8102.0) &  1028.5 & 8031 (8138.3) & 1.68 & \textbf{8023} & 13.08 & \textbf{8023}  & 3027  \\
r21 & 281  & 9284 & \textbf{9284} (9384.5) &  4588.7 & 9294 (9320.1) & 0.01 & \textbf{9284} & 9.15 & \textbf{9284}  & 6103  \\
r22 & 285  & 8887 & \textbf{8887} (8959.3) &  12911 & 8924 (9030.6) & 0.01 & \textbf{8887} & 63.42 & \textbf{8887}  & 1521  \\
r23 & 288  & 9136  & \textbf{9136} (9267.9) &  3252.0 &  9145 (9222.0) & 0.05 & \textbf{9136} & 42.77 & \textbf{9136} (9137.7) & 25716  \\
r24 & 269  & 8464 & \textbf{8464} (8572.9) &  13142.6 & 8468 (8534.5) & 0.01 & \textbf{8464} & 0.51 & \textbf{8464}  & 797  \\
r25 & 266  & 8426 & 8468 (8560.8) &  874.8 & 8579 (8649.6) & 0.01 & \textbf{8426} & 36.03 & \textbf{8426}  & 7088  \\
r26 & 284  & 8819 & \textbf{8819} (8927.9) &  14225.1 & 8937 (9035.3) & 0.01 &  \textbf{8819} & 82.2 & \textbf{8819}  & 22861  \\
r27 & 259  & 7975 & \textbf{7975} (8019.7) &  14074.9 & \textbf{7975} (7997.3) & 1.71 & \textbf{7975} & 9.86 & \textbf{7975} & 102  \\
r28 & 288  & 9407 & \textbf{9407} (9599.4) &  8691.0 & 9409 (9475.4) & 0.01 & \textbf{9407} & 0.44 & \textbf{9407}  & 1891  \\
r29 & 281  & 8693 & \textbf{8693} (8743.7) &  7613.1 &  8701 (8743.7) &  0.03 & \textbf{8693} & 4.54 & \textbf{8693} & 3429  \\
r30 & 301  & 9816  & \textbf{9816} (10003.2) &  8838.6 & 9820 (9877.1) & 0.01 & \textbf{9816} & 1.36 & \textbf{9816}  & 147  \\
\hline
 & & & &  & & & & & &  \\
 Best rate & & & & 87\% & & 13.3\% & & 100 \% & & 100\% \\
\hline
\end{tabular}

\end{table*}

\begin{table*}[!h]
\centering
\tiny
\caption{Comparative results of DLMCOL  with the state-of-the-art methods (MWSS, AFISA, RedLS, ILS-TS) for DIMACS/COLOR large instances of the WVCP. Dominating  results are indicated in boldface. New upper bounds are displayed with a star.  \label{results4}}
\begin{tabular}{lll|ll|ll|ll|ll}
   \hline
 \multicolumn{3}{c|}{Instance}   & \multicolumn{2}{|c|}{AFISA} & \multicolumn{2}{|c|}{RedLS} & \multicolumn{2}{|c|}{ILS-TS} & \multicolumn{2}{|c}{DLMCOL} \\
   \hline
  Graph name & $|V|$ & Opti. & Best (Avg.) & t (s) & Best (Avg.) & t (s) & Best (Avg.) & t (s)  &  Best (Avg.) & t (s)   \\
    \hline
C2000.5 & 2000  & - & 2400 (2425.1) &  3134.0 & \textbf{2151} (2162.4) & 11827 & 2250 (2266.2) & 11030 & - & - \\
 C2000.9 & 2000 & - & 6228 (6284.0) &  2798.3 & \textbf{5486} (5507.9) & 166246 & 5808 (5849.3) & 161980 & - & - \\    
 DSJC1000.1 & 1000  & - & 359 (362.9) &  430.5 & \textbf{300} (302.6) & 98115 & 305 (305.9) & 97025 & 342.0 (344.5)  & 1105 \\
 DSJC1000.5 & 1000   & - & 1357 (1430.9) &  371.7 & \textbf{1220} (1228.4) & 234 & 1242 (1270.4) & 1929 & 1230.0 (1260.0) & 167639 \\
DSJC1000.9 & 1000  & - & 3166 (3231.0) & 490.2 & \textbf{2864} (2875.7) & 48298 & 2975 (2997.8) & 101238 & - & - \\
 DSJC250.1 & 250  & - & 140 (141.9) & 48.9 & 130 (132) & 1 & \textbf{127} (127) & 2576 & \textbf{127} (127) & 1353 \\
 DSJC250.5 & 250  & -  & 415 (428.1) & 269.2 & 404 (407.7) & 43579 & 393 (393.3) & 58615 & \textbf{392} & 9226 \\
 DSJC250.9 & 250  & 934 & 939 (943.2) & 926 & 940 (943.11) & 801.22 & \textbf{934} (936.4) & 7670.05 & \textbf{934} & 4722 \\
 DSJC500.1 & 500  & - & 210 (215.6) & 426 & 188 (189.8) & 50 & \textbf{184} (185.4) & 72845 & \textbf{184} (184.9) & 23049 \\
DSJC500.5 & 500  & - & 778 (845.1) & 159.3 & 726 (729.0) & 2466 & 707 (716.9) & 51407 & \textbf{685}* (688.4) & 47064 \\
DSJC500.9 & 500  & - & 1790 (1854.5) & 831.1 & 1681 (1685.3) & 148914 & 1692 (1702.5) & 136637 & \textbf{1662}* (1664) & 121518 \\
DSJR500.1 & 500  & - & \textbf{169} (175.4) & 458.9 & 171 (173.4)  & 1 & \textbf{169} (169) & 0.56 & \textbf{169} (171.7) & 35389 \\
flat1000\_50\_0 & 1000  & - & 1289 (1315.7) & 981.8 & 1184 (1186.7)  & 93711 & 1218 (1226.4) & 141135 & \textbf{924}*  & 82068 \\
flat1000\_60\_0 & 1000  & - & 1338 (1354) & 201.9 & \textbf{1220} (1230.6)  & 264 & 1242 (1258.0) & 60631 & 1224 (1231.6) & 168937 \\
flat1000\_76\_0 & 1000  & - & 1314 (1337.6 ) & 2396.6 & \textbf{1200} (1210.5)  & 316 & 1240 (1246.7) & 141292 & 1210.0 (1220.6) & 164934\\
inithx.i.1 & 864 & - & 587 (587.9 ) & 527.5 & \textbf{569} (571.4)  & 2 & \textbf{569}  & < 0.01 & \textbf{569}  & 214 \\
inithx.i.2 & 645  & - & 341 (341.6) & 0.01  & \textbf{329} (331.5)  & 1328 & \textbf{329}  & 184 & 336 (337.0) & 127 \\
inithx.i.3 & 621  & - & 352 (355.6 ) & 0.01 & \textbf{337} (337.4)  & 5 & \textbf{337}  & 1 & 350 (352.2) & 142 \\
latin\_square\_10 & 900  & - & 1690 (1900.0) & 780.3 & 1548 (1561.8) & 36451 & 1542 (1557.0) & 142925 & \textbf{1480}* (1486.2) &149656 \\
le450\_15a & 450  & - & 241 (247.1) & 288.4 & 217 (218.7) & 1 & \textbf{213} (213.8) & 83054 & \textbf{213} (216.5) & 43854 \\
le450\_15b & 450  & - & 239(245.1) & 368.3 & 219 (225.3) & 1 & \textbf{217} (218.2) & 2505 & \textbf{217} (219.6) & 71563 \\
le450\_15c & 450  & - & 313 (320.8) & 432.9 & 288 (292.2) & 2 & 277 (280.3) & 17392 & \textbf{275}* (277.4)  & 46789\\
le450\_15d & 450  & - & 306 (314.1) & 113.7 & 285 (288.9) & 979 & 274 (275.8) & 155735 & \textbf{272}* (272.4) & 22917\\
le450\_25a & 450  & - & 317 (329.9) & 362.3 & 308 (310.4) & 45732 & \textbf{306} (306.0) & 715 & 307 (308.6) & 26228\\
 le450\_25b & 450 & - & 318 (325.8) & 285.9 & 308 (310.4) & 45732 & \textbf{307} (307.0) & 18 & \textbf{307} (308.8) & 43456 \\
 le450\_25c & 450  & - & 378 (387.9 ) & 359.4 & 360 (364.1) & 2 & 349 (352.1)& 82371 & \textbf{342}* (343) & 57924\\
 le450\_25d & 450  & - & 375 (385.3) & 254.8 & 342 (350.2) & 4 & 339 (342.4) & 16881 & \textbf{330}* (330.1) & 45128\\
 miles1000 & 128  & 431 & 432 (444.7 ) & 480.0 & \textbf{431} & 6.06 & \textbf{431} & 18.81 & \textbf{431} & 581 \\
 miles1500 & 128  & 797 & \textbf{797}  & 1802 & \textbf{797} (797.3) & 0.03 & \textbf{797} & 1.68 & \textbf{797} & 76 \\ 
 miles250 & 128  & 102 & \textbf{102} (102.7) & 56.6 & \textbf{102} (103.6) & 0.7 & \textbf{102} & 0.18 & \textbf{102} & 21 \\
 miles500 & 128  & 260 & \textbf{260} (261.3) & 48.4 &  \textbf{260} (260.4) & 1.39 & \textbf{260} & 0.18 & \textbf{260} & 30 \\
 queen10\_10 & 100  & - & 166 (169.2 ) & 68.4 & \textbf{162} (165.3) & 10400 & \textbf{162} (162.0) & 24 & \textbf{162} (162) & 19 \\
 queen11\_11 & 121  & - & 178 (182.3) & 55.2 & 179 (180.1)  & 909 & \textbf{172} (172.7) & 68208 & \textbf{172} (172) & 1753 \\
 queen12\_12 & 144  & - & 194 (198.6) & 92.7 & 188 (189.4)  & 133726 & \textbf{185} (185.2) & 23709 & \textbf{185} (185) & 1826 \\
 queen13\_13 & 169  & - & 204 (207.5) & 199.8 & 197 (200.9) & 8100 & \textbf{194} (194.8) & 20802 & \textbf{194} (194) & 1150 \\
 queen14\_14 & 196  & - & 224 (227.4) & 360.1 & 217 (219.7)  & 2 & 216 (217.1) & 469 & \textbf{215}* (215.2) & 16621 \\
 queen15\_15 & 225  & - & 237 (241.2) & 183.4 & 230 (233.5) & 1 & 224 (226.7) & 2685 & \textbf{223}* (224.1) & 9276\\
 queen16\_16 & 256  & - & 253 (256.3) & 300.9 & 242 (246.0)  & 579 & 238 (239.0) & 60187 & \textbf{234}* (234.8) & 14751  \\
 wap01a & 2368  & -& 638 (653.1) & 1133.5 & \textbf{545} (558.6)  & 89184 & 549 (550.5) & 62497 & - &  - \\
 wap02a & 2464  & - & 637 (638.1) & 3270.46 & \textbf{538} (547.1)  & 40143 & 541 (543.1) & 46848 & - & -  \\ 
wap03a & 4730  & - & 687 (707.5) & 2901.5 & \textbf{562} (566.9)  & 134923 & 577 (579.7) & 55554 & - & -  \\ 
wap04a & 5231  & - & 698 (709.0 ) & 4.79 & \textbf{563} (583.0)  & 52833 & 570 (573.0) & 538455 & - & -  \\  
wap05a & 905  & - & 598 (610.9) & 1574.5 & \textbf{542} (548.1)  & 1348 & \textbf{542} (542.8) & 62323 & 587 (588.0) & 1139  \\  
wap06a & 947  & - & 599  (607.6) & 65.3 & \textbf{519} (529.4)  & 134 & \textbf{519} (520.8) & 46077 & 574 (587.0) & 1244  \\  
wap07a & 1809  & - & 680 (692.5) & 384.82 & \textbf{561} (563.8)  & 3433 & 567 (571.0) & 11943 & 685 (686.0) & 8294  \\  
wap08a & 1870  & - & 663 (673.4) & 2627.2 & \textbf{529} (540.3)  & 37505 & 546 (549.7) & 67847 & 663 (665) & 7397  \\ 
zeroin.i.1 & 211  & 511 & 518  & 0.01 & \textbf{511}  & 0.08 & \textbf{511}  & 1.7 & \textbf{511} & 38  \\  
zeroin.i.2 & 211  & 336 & \textbf{336} (337.6) & 440.8 & \textbf{336}  & 0.22 & \textbf{336}  & 0.01 & \textbf{336} & 33  \\  
zeroin.i.3 & 206  & 298 & 299 (301.7)  & 139.6 & \textbf{298} (298.7) & 2.25 & \textbf{298}  & 10.67 & \textbf{298} & 31  \\  
\hline
 & & & &   & & & & &  \\
 Best rate &  & & 10.2\% & & 53.1\% & & 48.9 \% & & 63.3\% \\
\hline
\end{tabular}

\end{table*}

\subsection{Comparative results on vertex coloring benchmark }

We show in this section the generality of the proposed approach by applying the approach to the vertex coloring problem. We present computational results on the 18 difficult DIMACS instances for the COL with respect to 12 state-of-the-art graph coloring methods. These instances are challenging because only a few algorithms can reach the best known results shown in Table \ref{table:gcp_results1} and only very few algorithms can attain the best known results for five graphs DSJC500.5, DSJC1000.5, flat\_300\_28\_0, flat\_1000\_76\_0, latin\_sqr\_10. Given that the COL is a special case of the WVCP when the vertex weight is equal to one, we also tested, for the first time, the two best WVCP algorithms,  RedLS \cite{wang2020reduction} and ILS-TS \cite{nogueira2021iterated}, on these difficult DIMACS instances\footnote{AFISA \cite{sun2017feasible} was not launched as the available binary code does not allow setting a cutoff time}. Each algorithm was run 10 times to solve each instance under the relaxed condition of 48 hours per instance and per run. We observe that these local search based WVCP algorithms perform similarly compared to other popular local search coloring algorithms (see Table \ref{table:gcp_results2} below).

Table \ref{table:gcp_results1} summarizes the computational results of our DLMCOL algorithm.  Columns 2 and 3 give the features of the tested instance: the number of vertices $|V|$ and the density of the graph (dens.). Columns 4-6 present the chromatic number of the graph ($\chi(G)$, when known) and  the best upper bound ($k^*$) reported so far in the literature with the references, including the two WVCP algorithms (RedLS \cite{wang2020reduction} and ILS-TS \cite{nogueira2021iterated}). In columns 7-9, the computational statistics of our DLMCOL algorithm are given, with the best (smallest) number of colors obtained to reach a legal solution ($k_{best}$), the associated success rate (SR) and the average time in seconds to reach the solution with the given $k$. 

Following the common practice to report comparative results in the coloring literature, we also display in Table \ref{table:gcp_results2}  the best solution found by each algorithm corresponding to the smallest number $k$ of colors needed to reach a legal coloring for a graph. Column 2 corresponds to the best $k$ found in the literature. Columns 3 displays the best $k$ found by DLMCOL. Columns 4-9 and columns 10-17 respectively report the best $k$ found by state-of-the art local search algorithms, including the two WVCP algorithms (RedLS \cite{wang2020reduction} and ILS-TS \cite{nogueira2021iterated}), and population based algorithms. Results corresponding to the best $k$ found so far are displayed in boldface. 

In Table \ref{table:gcp_results2}, we regroup the three references \cite{titiloye2011graph}, \cite{titiloye2011quantum} and \cite{titiloye2012parameter}  in the same column (Column 16) as they correspond to the same QA-COL algorithm launched with different parameters and using different computing tools. We report the best $k$ jointly reported in these three references. 

It should be mentioned that these DIMACS instances have been studied for a long time (over thirty years) and some of the best known results have only been obtained by very few algorithms and sometimes with a very low success rate.  Different computing tools have been used (such as multiple core servers with parallel computing) under specific and relaxed conditions (e.g., large run time from several days to one month, specific fine tuning of the hyperparameters for each given instance, etc.). 

As displayed in Tables \ref{table:gcp_results1} and \ref{table:gcp_results2}, DLMCOL can reach all the best results  in the literature for these instances except for latin\_sqr\_10, for which a solution with $k=97$ was only found once in \cite{titiloye2011quantum}. In general, DLMCOL is very competitive for solving very difficult instances of medium size such as DSJC500.5 and flat300\_28\_0.  Notably it finds a legal 47-coloring for DSJC500.5 with a success rate of 5/10 (only two reference algorithms QA-COL \cite{titiloye2012parameter} and HEAD  \cite{moalic2018variations} can reach this result occasionally with specifically fine-tuned hyperparameters).  DLMCOL also finds a solution with $k=28$ colors for flat300\_28\_0, which is difficult for the two best competitors QA-COL \cite{titiloye2011graph,titiloye2011quantum,titiloye2012parameter} and HEAD \cite{moalic2018variations}. For this instance, performing a strong exploration of the search space with a large population is very beneficial, as it seems that there exists only one legal solution with $k=28$ colors. Indeed, dozens of solutions found by our algorithm are always the same up to color permutation. Finally, one notices that the DLMCOL algorithm is quite time consuming to solve large instances, given that it uses a very large population.

 %\begin{landscape}
\begin{table*}[!h]
\centering
\caption{Computational results of DLMCOL on the \textit{difficult} DIMACS coloring challenge benchmarks for the COL problem}.
\footnotesize
\begin{tabular}{l|ll|lll|lll} 
   \hline
   & \multicolumn{5}{c}{}  &  \multicolumn{3}{|c}{DLMCOL}\\
   \hline
  Instance & $|V|$ & dens. & $\chi(G)$ & $k^*$ &  references & $k_{\text{best}}$ & SR & t(s) \\
    \hline
    DSJC250.5 & 250 & 0.5 & ? & 28  & \cite{blochliger2008graph,galinier1999hybrid,galinier2008adaptive,goudet2021population,hertz2008variable,lu2010memetic,malaguti2008metaheuristic,moalic2018variations,porumbel2009diversity,titiloye2011quantum,nogueira2021iterated,wang2020reduction} &  \textbf{28} & 10/10  & 219\\
    DSJC500.1 & 500 & 0.1 & ? & 12  & \cite{blochliger2008graph,chiarandini2002application,dorne1999tabu,galinier1999hybrid,galinier2008adaptive,goudet2021population,hertz2008variable,lu2010memetic,malaguti2008metaheuristic,moalic2018variations,porumbel2009diversity,titiloye2011quantum} &  \textbf{12} & 10/10  & 2396\\
    DSJC500.5 & 500 & 0.5 & ? & 47  & \cite{moalic2018variations,titiloye2012parameter} &  \textbf{47} & 5/10  & 42513\\    
    DSJC500.9 & 500 & 0.9 & ? & 126  & \cite{chiarandini2002application,galinier2008adaptive,goudet2021population,hertz2008variable,lu2010memetic,moalic2018variations,porumbel2009diversity,titiloye2011quantum,wang2020reduction} &  \textbf{126} & 10/10  & 17172\\   
    DSJC1000.1 & 1000 & 0.1 & ? & 20  & \cite{blochliger2008graph,galinier1999hybrid,galinier2008adaptive,goudet2021population,hertz2008variable,lu2010memetic,malaguti2008metaheuristic,moalic2018variations,porumbel2009diversity,titiloye2011quantum} &  \textbf{20} & 10/10  & 32477\\ 
    DSJC1000.5 & 1000 & 0.5 & ? & 82  & \cite{moalic2018variations,titiloye2012parameter} &  \textbf{82} & 4/10  & 167038\\ 
    DSJR500.1c & 500 & 0.97 & 85 & 85  & \cite{blochliger2008graph,galinier1999hybrid,goudet2021population,hertz2008variable,malaguti2008metaheuristic,moalic2018variations,nogueira2021iterated,porumbel2009diversity,titiloye2011quantum,wang2020reduction} &  \textbf{85} & 10/10  & 4395\\  
    DSJR500.5 & 500 & 0.47 & 122 & 122  & \cite{goudet2021population,lu2010memetic,malaguti2008metaheuristic,moalic2018variations,nogueira2021iterated,prestwich2002coloration,titiloye2011quantum,wang2020reduction} &  \textbf{122} & 10/10  & 4310\\ 
   r250.5 & 250 & 0.48 & 65 & 65  & \cite{goudet2021population,lu2010memetic,malaguti2008metaheuristic,moalic2018variations,nogueira2021iterated,titiloye2011quantum,nogueira2021iterated,wang2020reduction} &  \textbf{65} & 10/10  & 19239\\ 
     r1000.1c & 1000 & 0.97 & ? & 98  & \cite{blochliger2008graph,dorne1999tabu,goudet2021population,lu2010memetic,malaguti2008metaheuristic,moalic2018variations,nogueira2021iterated,porumbel2009diversity,titiloye2011quantum,wang2020reduction} &  \textbf{98} & 10/10  & 2102\\
    le450\_25c & 450 & 0.17 & 25 & 25  & \cite{blochliger2008graph,goudet2021population,hertz2008variable,lu2010memetic,malaguti2008metaheuristic,moalic2018variations,porumbel2009diversity,titiloye2011quantum} & \textbf{25} & 10/10  & 48201\\   
    le450\_25d & 450 & 0.17 & 25 & 25  & \cite{blochliger2008graph,goudet2021population,hertz2008variable,lu2010memetic,malaguti2008metaheuristic,moalic2018variations,porumbel2009diversity,titiloye2011quantum} & \textbf{25} & 10/10  & 45038\\ 
    flat\_300\_26\_0 & 300 & 0.48 & 26 & 26  & \cite{chiarandini2002application,dorne1999tabu,goudet2021population,lu2010memetic,malaguti2008metaheuristic,nogueira2021iterated,titiloye2011quantum,wang2020reduction} & \textbf{26} & 10/10  & 113\\ 
    flat\_300\_28\_0 & 450 & 0.17 & 28 & 28  & \cite{blochliger2008graph,hertz2008variable} & \textbf{28} & 4/10  & 24651\\     
    flat\_1000\_50\_0 & 1000 & 0.49 & 50 & 50  & \cite{blochliger2008graph,dorne1999tabu,galinier2008adaptive,hertz2008variable,lu2010memetic,malaguti2008metaheuristic,moalic2018variations,nogueira2021iterated,porumbel2009diversity,titiloye2011quantum,wang2020reduction} & \textbf{50} & 10/10  & 7160 \\  
    flat\_1000\_60\_0 & 1000 & 0.49 & 60 & 60  & \cite{dorne1999tabu,blochliger2008graph,galinier2008adaptive,hertz2008variable,lu2010memetic,malaguti2008metaheuristic,moalic2018variations,porumbel2009diversity,titiloye2011quantum,wang2020reduction} & \textbf{60} & 10/10  & 12110 \\ 
    flat\_1000\_76\_0 & 1000 & 0.49 & 76 & 81  & \cite{moalic2018variations,titiloye2012parameter} & \textbf{81} & 3/10  & 249165 \\  
    latin\_sqr\_10 & 900 & 0.76 & ? & 97  & \cite{titiloye2011graph} & 98 & 8/10  & 170518 \\   
    \hline
\end{tabular}
\label{table:gcp_results1}
\end{table*}

\begin{table*}[!h]
\centering
\footnotesize
\caption{Comparison of DMLCOL with the state-of-the-art algorithms in terms of the best results on the \textit{difficult} DIMACS coloring challenge benchmarks for the COL problem}.
\begin{tabular}{l|l|l|llllll|lllllllll} 
   \hline
   & &  & \multicolumn{6}{c|}{Local search algorithms} &  \multicolumn{8}{c}{Population based algorithms}\\
   \hline
  Instance & $k^*$ &  DLMCOL & \cite{dorne1999tabu} & \cite{chiarandini2002application} & \cite{hertz2008variable} & \cite{blochliger2008graph}  & \cite{wang2020reduction} &\cite{nogueira2021iterated} & \cite{galinier1999hybrid} & \cite{galinier2008adaptive} & \cite{malaguti2008metaheuristic} & \cite{porumbel2009diversity} & \cite{lu2010memetic} & [42-44]  & \cite{moalic2018variations} & \cite{goudet2021population} \\
    \hline
    DSJC250.5 & 28 & \textbf{28} & - & - & \textbf{28} & \textbf{28} & \textbf{28} & \textbf{28} & \textbf{28} & \textbf{28} & \textbf{28} & \textbf{28} & \textbf{28} & \textbf{28}  & \textbf{28} & \textbf{28}\\
    DSJC500.1 & 12  & \textbf{12} & \textbf{12} & \textbf{12} & \textbf{12} & \textbf{12}  & 13 & 13 & \textbf{12} & \textbf{12} & \textbf{12} & \textbf{12} & \textbf{12}& \textbf{12}  & \textbf{12} & \textbf{12}\\
    DSJC500.5 & 47 & \textbf{47} & 50 & 49 & 48 & 48  & 50 & 50 & 48 & 48 & 48 & 48 & 48 &  \textbf{47} & \textbf{47} & 48  \\    
    DSJC500.9 & 126 & \textbf{126} & 127 & \textbf{126} & \textbf{126} & 127  & \textbf{126} & 127 & - & \textbf{126} & 127 & \textbf{126} & \textbf{126} & \textbf{126} & \textbf{126} & \textbf{126}\\   
    DSJC1000.1 & 20 & \textbf{20} & 21 & - & \textbf{20} & \textbf{20}  & 21 & 21 & \textbf{20} & \textbf{20} & \textbf{20} & \textbf{20} & \textbf{20} & \textbf{20}  & \textbf{20} & \textbf{20} \\ 
    DSJC1000.5 & 82 & \textbf{82} & 90 & 89 & 86 & 89 & 91 & 91 & 83 & 84 & 83 & 83 & 83 &  \textbf{82} & \textbf{82} & 84\\ 
    %  DSJC1000.9 & 222 & - & 226 & - & 224 & 226 & 224 & 224 & 224 & 224 & 223 & 222 & 222 & 222 & 224\\   
    DSJR500.1c & 85 & \textbf{85} & - & - & \textbf{85} & \textbf{85}  &  \textbf{85} &  \textbf{85} & \textbf{85} & 86 & \textbf{85} & \textbf{85} & - & \textbf{85}  & \textbf{85} & \textbf{85}\\  
    DSJR500.5 & 122 & \textbf{122} & - & 124 & 125 & 125  & \textbf{122} & \textbf{122} & - & 127 & \textbf{122} & 124 & \textbf{122} & \textbf{122} & \textbf{122} & \textbf{122}\\ 
   r250.5 & 65 & \textbf{65} & - & - & 66 & 66 & \textbf{65} & \textbf{65} & - & - & \textbf{65} & - & \textbf{65} & \textbf{65}  & \textbf{65} & \textbf{65} \\ 
     r1000.1c & 98 & \textbf{98} & \textbf{98} & - & - & \textbf{98}  & \textbf{98} & \textbf{98} & - & - & \textbf{98} & \textbf{98} & \textbf{98} & \textbf{98}  & \textbf{98} & \textbf{98}\\
    % r1000.5 & 234 & - & 242 & - & - & 248 & - & - & 234 & 245 & 245 & 238 & - & 245 & 234  \\ 
    le450\_25c & 25 & \textbf{25} & - & 26 & \textbf{25} & \textbf{25}  & 26 & 26 & 26 & 26 & \textbf{25} & \textbf{25} & \textbf{25} & \textbf{25}  & \textbf{25} & \textbf{25}\\   
    le450\_25d & 25 & \textbf{25} & - & 26 & \textbf{25} & \textbf{25}  & 26 & 26 & - & 26 & \textbf{25} & \textbf{25} & \textbf{25} & \textbf{25}  & \textbf{25} & \textbf{25} \\ 
    flat\_300\_26\_0 & 26 & \textbf{26} & \textbf{26} & \textbf{26} & - & - & \textbf{26} & \textbf{26} & - & \textbf{26} & \textbf{26} & - & \textbf{26} & \textbf{26}  & \textbf{26} & \textbf{26}\\ 
    flat\_300\_28\_0 & 28 & \textbf{28} & 31 & 31 & \textbf{28} & \textbf{28}  & 31 & 32 & 31 & 31 & 31 & 31 & 29 & 31  & 31 & 31\\     
    flat\_1000\_50\_0 & 50 & \textbf{50} & \textbf{50}  & - & \textbf{50}  & \textbf{50}  & \textbf{50} & \textbf{50}   & - & \textbf{50} & \textbf{50} & \textbf{50} & \textbf{50} & \textbf{50}  & \textbf{50} & \textbf{50}\\  
    flat\_1000\_60\_0 &  60 & \textbf{60} & \textbf{60} & - & \textbf{60} & \textbf{60}  &  \textbf{60} & 90 & - & \textbf{60} & \textbf{60} & \textbf{60} & \textbf{60} & \textbf{60}  & \textbf{60} & \textbf{60}\\ 
    flat\_1000\_76\_0 & 81 & \textbf{81} & 89 & - & 85 & 87  & 89 & 91 & 83 & 84 & 82 & 82 & 82  & \textbf{81} & \textbf{81} & 83 \\   
    latin\_sqr\_10 & 97 & 98 & - & 99 & - & -  & 101 & 100 & - & 104 & 101 & - & 99 & \textbf{97}  & - & 98 \\   
    % C2000.5 & 145 & - & - & - & - & - & - & - & - & 151 & 148 & - & 145 & 148 & - \\ 
    % C4000.5 & 270 & - & - & - & - & - & - & - & - & - & 272 & - & 270 & 275 & -\\         
%   \hline
%   best found & & 17/18 & 5/18 & 3/18 & 8/18 & 9/18 & 4/18 & 7/18 & 12/18 & 10/18 & 12/18 & 17/18 & 16/18 & 13/18  \\
    \hline
\end{tabular}
\label{table:gcp_results2}
\end{table*}

\section{Analysis of important factors in the algorithm \label{sec:key_components}} 

In this section, we analyze the impacts of two important factors of the DLMCOL algorithm: (i) the very large population and (ii) the contributions of deep learning. 

\subsection{Sensitivity to the population size} 

\begin{figure}[!ht]
    \centering
    \includegraphics[width=0.45\textwidth]{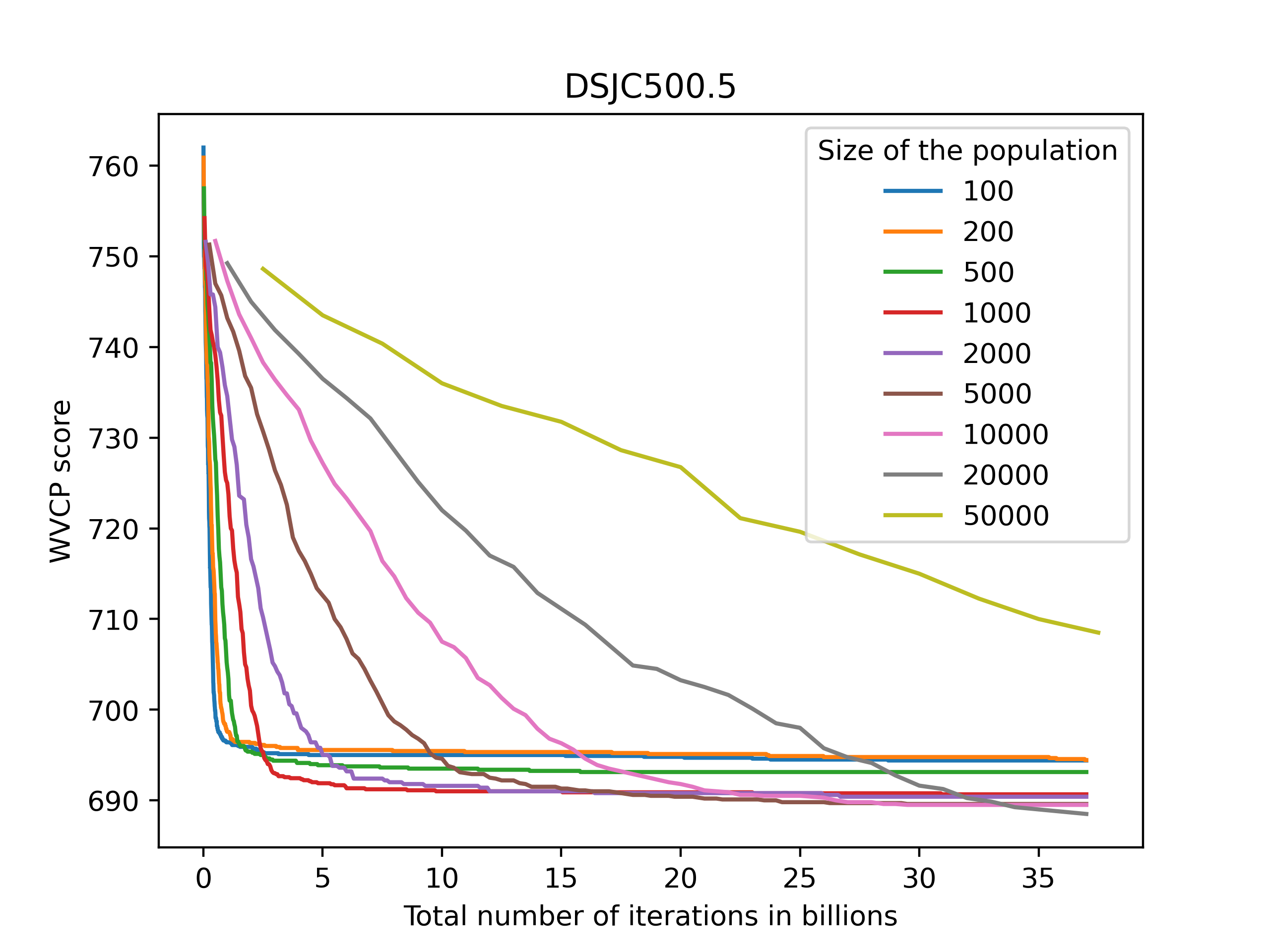}
    \caption{Impact of the population size $p$ on the performance of DLMCOL. $y$-axis corresponds to the WVCP score (average best score over 10 runs at each generation) and $x$-axis corresponds to the number of generations.}\label{fig:sensi_pop} 
\end{figure}

We perform a sensitivity analysis of the algorithm with respect to the population size $p$. For this, we launched the DLMCOL algorithm with $p$ taking nine different values in the range $[100,50000]$ to solve the instance DSJC500.5 of the WVCP with the same total number of tabu search iterations. Figure \ref{fig:sensi_pop} displays the sensitivity of the average results over 10 runs to the population size $p$. 

We observe that for the same total number of tabu search iterations, DLMCOL obtains better results with a large population size of $20000$ than with other sizes. This can be explained by two reasons. First, a large population improves the population diversity, which favors the finding of promising areas in the search space and helps to better train the neural network at each generation. Second, a large population increases the chance for each individual to find a closer but different neighbor in the population for parent matching, which helps to generate promising offspring solutions. However, an excessively large population (such as $p=50000$) is counterproductive, because the algorithm requires, in this case, much more time to converge toward solutions of good quality.

\subsection{Analysis of the contributions of deep learning} \label{sec:analysis}

We now analyze the interest of the neural network within the memetic framework. First, we study the general benefits of the crossover selection strategy driven by deep learning by performing an ablation study. Secondly, we compare the predicted results of local searches with the actual results, so as to shed lights on why this strategy is effective for some instances and less effective for others.

\paragraph{Benefits of the neural network based crossover selection} \label{sec:general_benefit}

To assess the contributions of the neural network driven crossover selection strategy (see Section \ref{Crossovers}), we launched 10 replications of DLMCOL on 4 instances (DSJC500.1, DSJC500.5, le450\_25c and le450\_25d) of the WVCP with a cutoff time of 20 hours, with and without the neural network crossover selection. In the version without neural network, the second parent of the crossover is randomly chosen among the $K$ nearest neighbors of each individual. 

\begin{figure*}[h]
    \centering
    \includegraphics[width=0.45\textwidth]{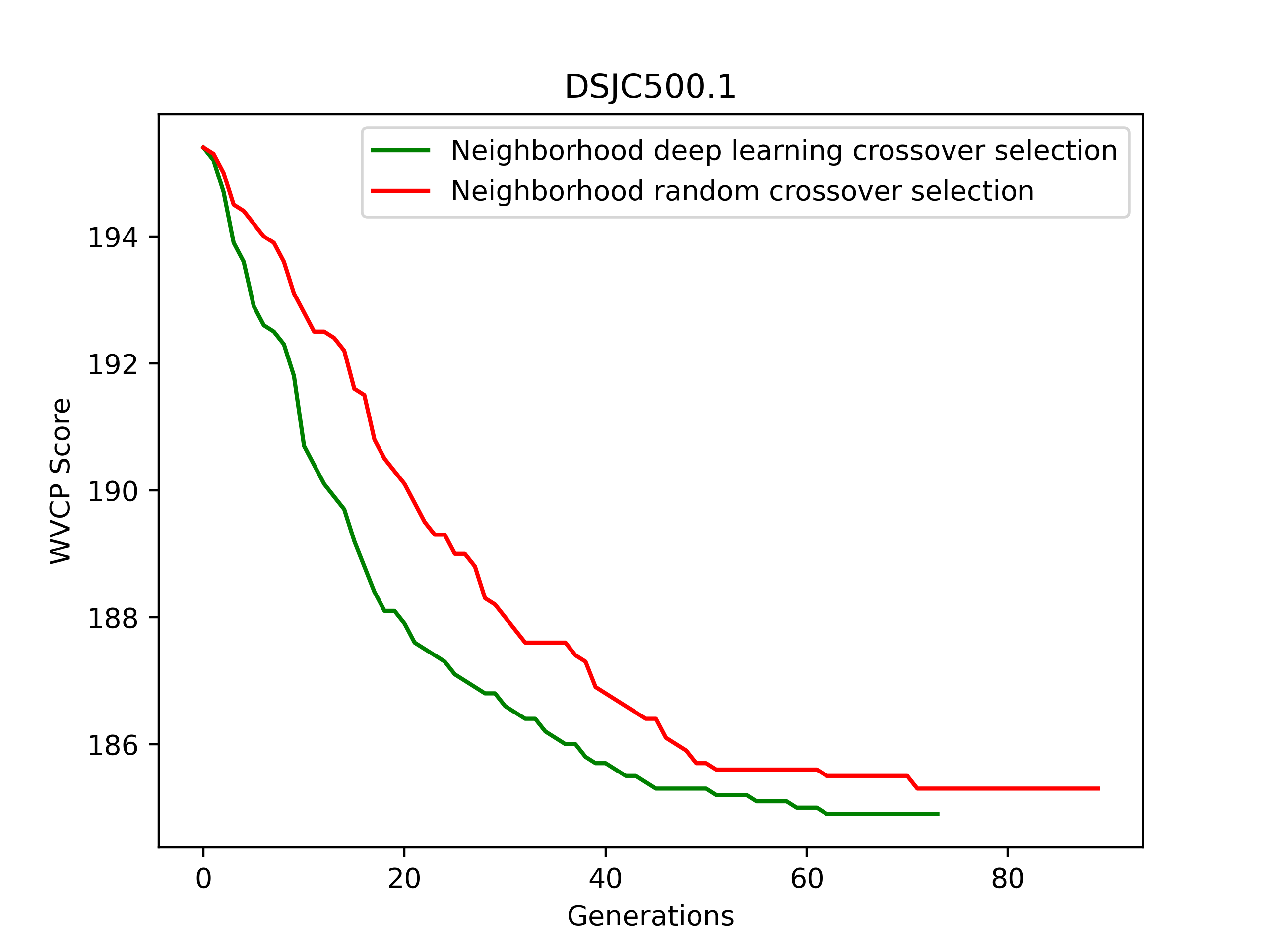}
    \includegraphics[width=0.45\textwidth]{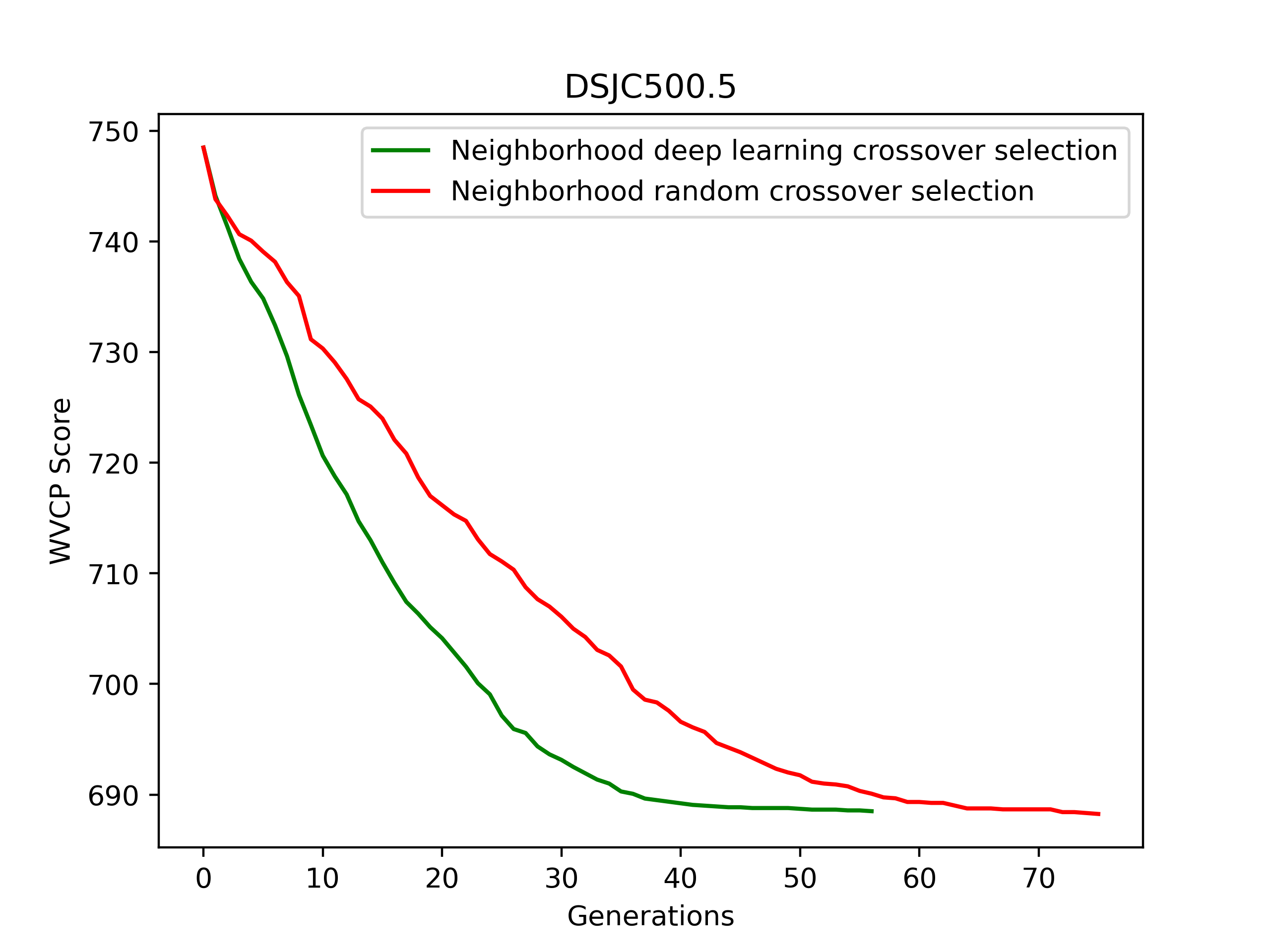}
    \includegraphics[width=0.45\textwidth]{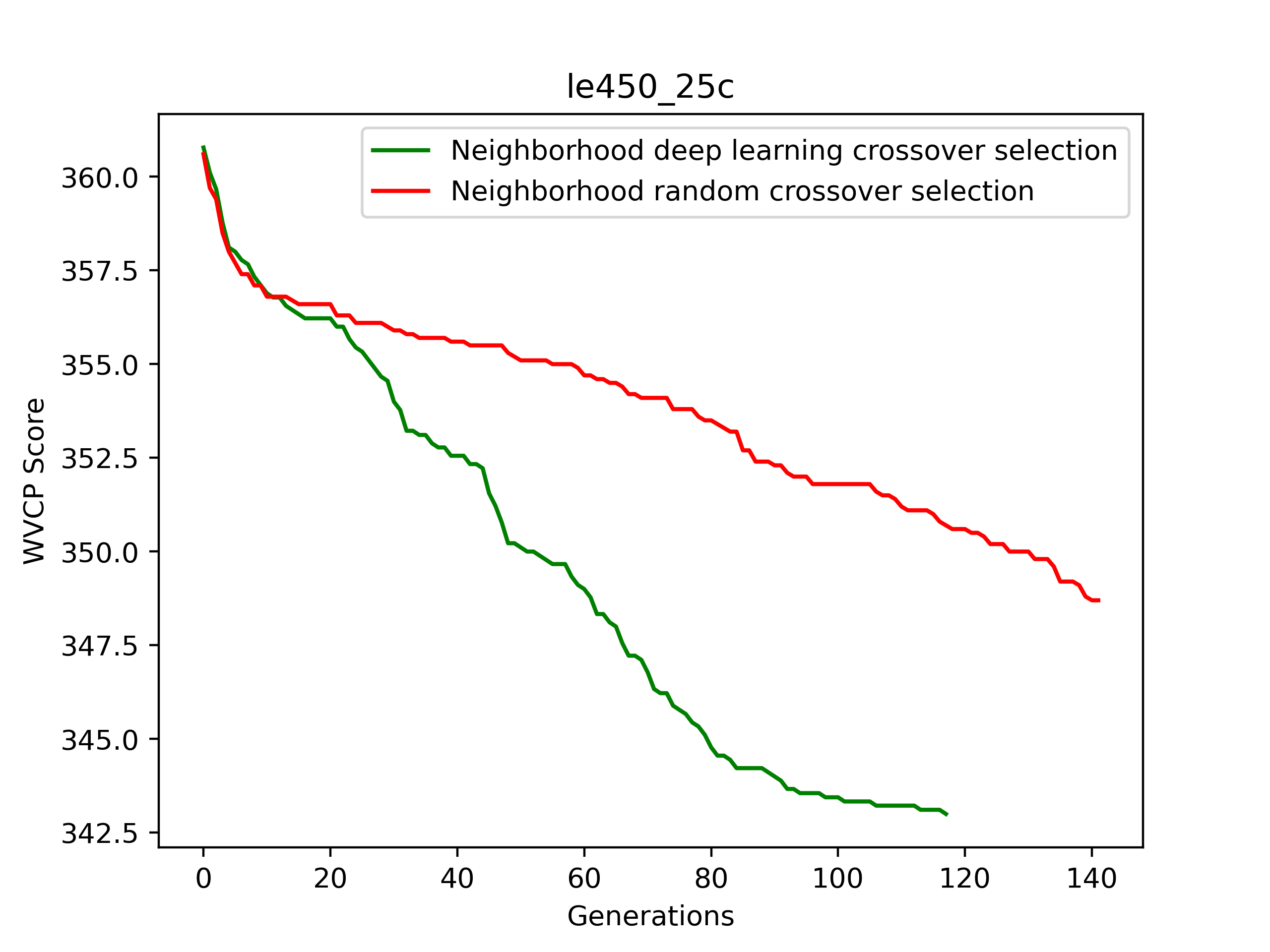}     
    \includegraphics[width=0.45\textwidth]{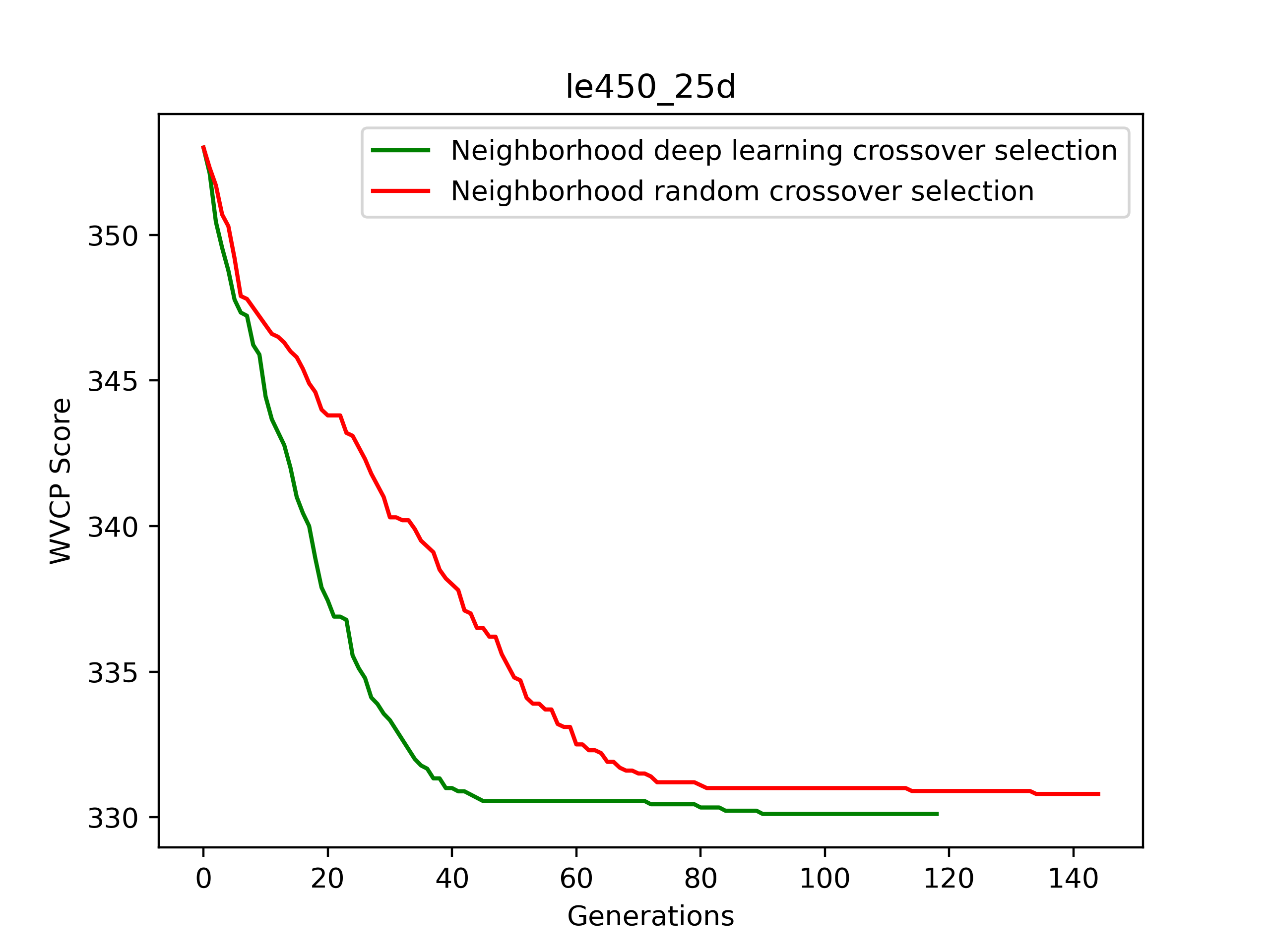} 
    \caption{Impact of the deep learning driven crossover selection strategy on the algorithm. $y$-axis corresponds to the WVCP score (average best score over 10 runs at each generation) and $x$-axis corresponds to the number of generations.}
    \label{fig:impactNN}
\end{figure*}

The average best score of the WVCP obtained at each generation is displayed on Figure \ref{fig:impactNN}. Green curve corresponds to the standard DLMCOL algorithm while the red curve corresponds to the version without deep learning. One first notices  that the version without deep learning can perform more generations in the same amount of time because no time is spent on the neural network training and offspring evaluations. On the other hand, we observe that the green curve is always below the red curve and that better results are achieved in the same amount of time. This highlights that the neural network can really contribute to a better selection of promising  crossovers for the memetic algorithm. 

\paragraph{Quality of the predictions given by the neural network}\label{sec:analysis_expected_fit}

Once trained, the neural network is used  in DLMCOL to predict in advance the results of local searches in order to select the best promising crossovers for the next generation (see Section \ref{Crossovers}). Therefore, one can wonder if the   fitness values predicted by the neural network for the $p$ new starting points, $\{f_{\theta}(S^O_1),\dots,f_{\theta}(S^O_p)\}$ at generation $t$ are close to the actual fitness values $\{f(S'_1),\dots,f_{\theta}(S'_p\}$ obtained after the $p$ local searches  at the next generation $t+1$. 

\begin{figure*}[!h]
    \centering
    \includegraphics[width=0.45\textwidth]{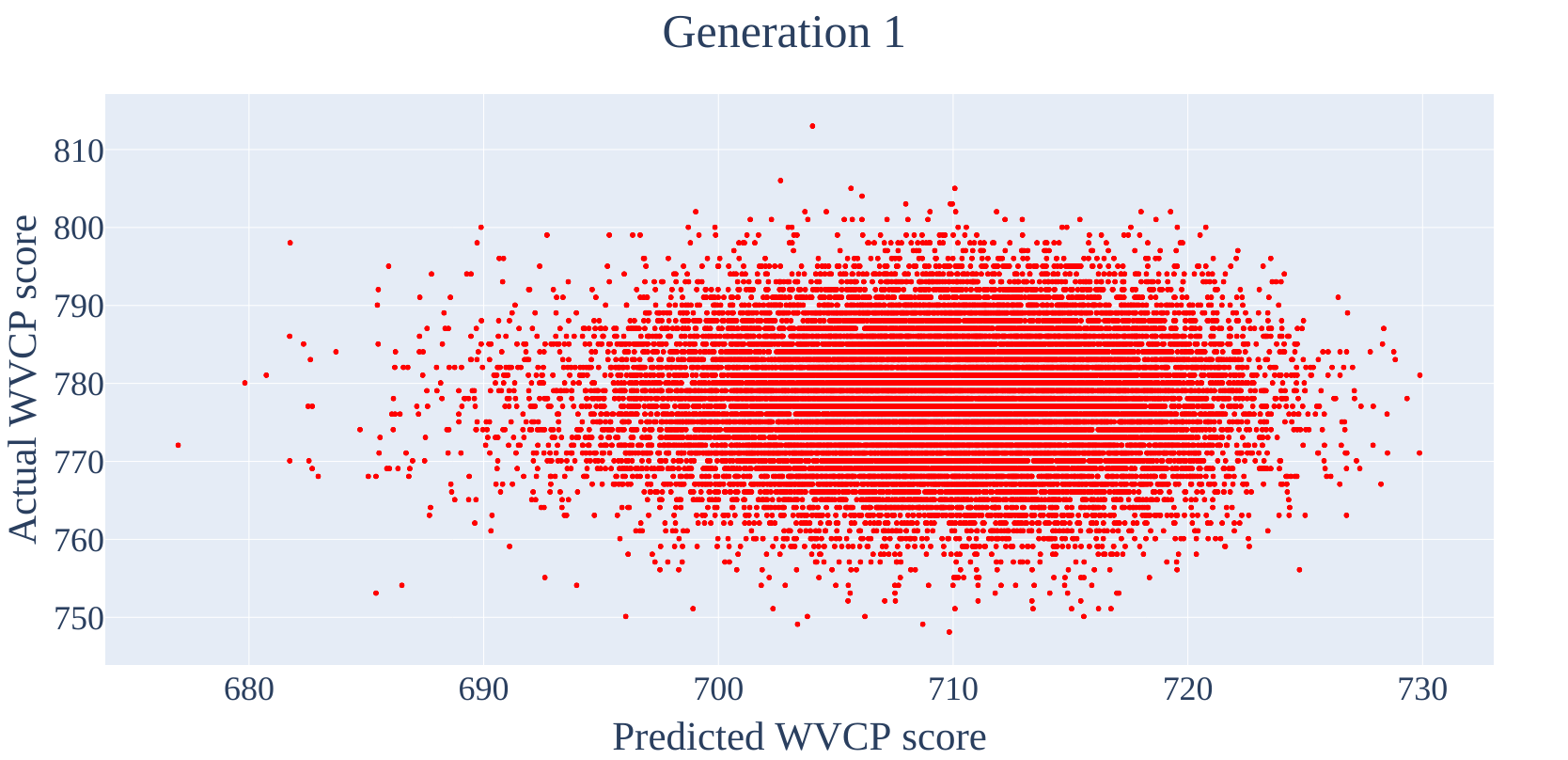}
    \includegraphics[width=0.45\textwidth]{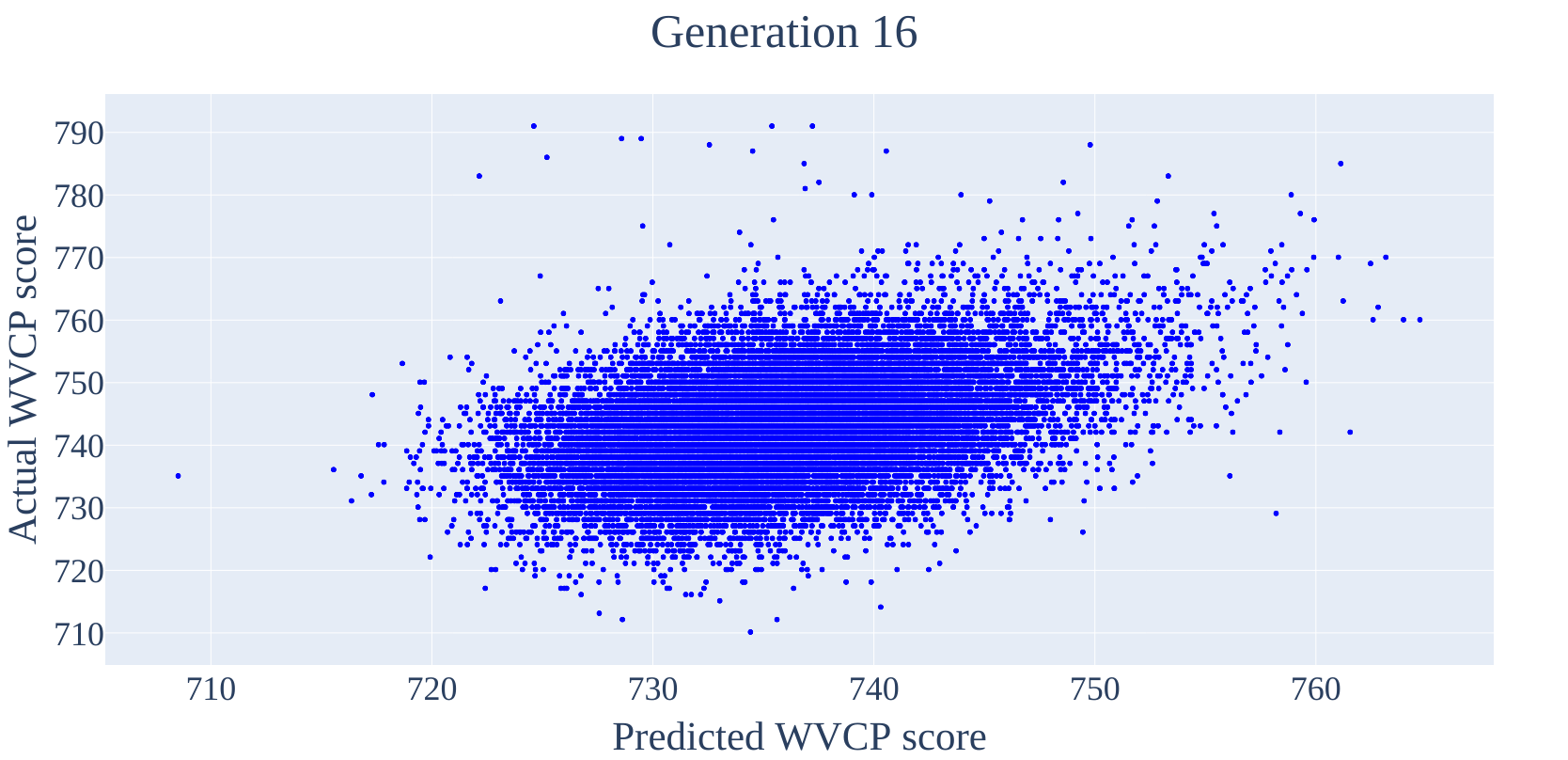}
    \includegraphics[width=0.45\textwidth]{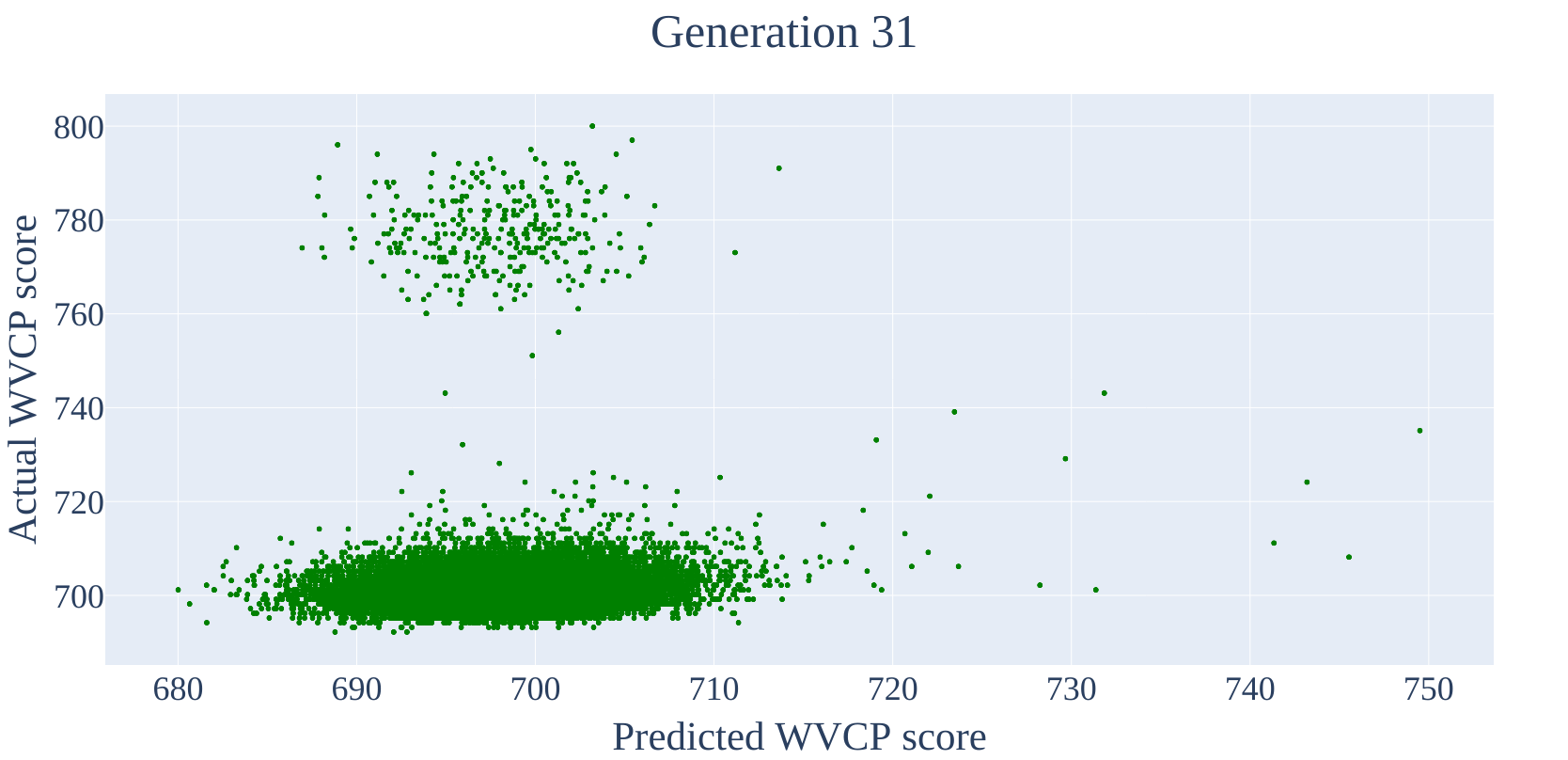}     
    \includegraphics[width=0.45\textwidth]{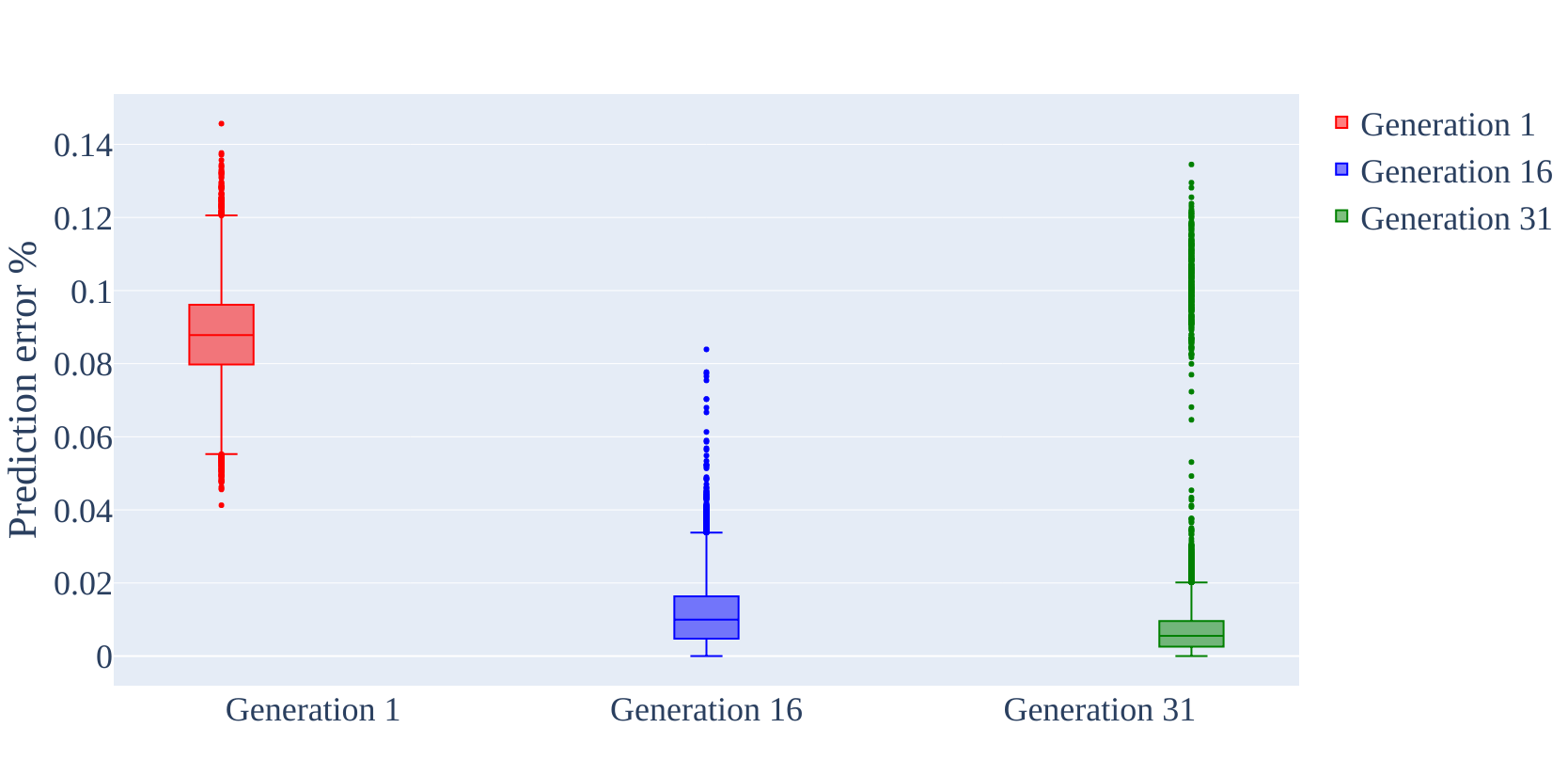} 
    \caption{Comparison of the predicted and actual results of the local searches at generation 1, 16 and 31 for the instance DSJC500.5.col.}
    \label{fig:expected_results1}
\end{figure*}

\begin{figure*}[!h]
    \centering
    \includegraphics[width=0.45\textwidth]{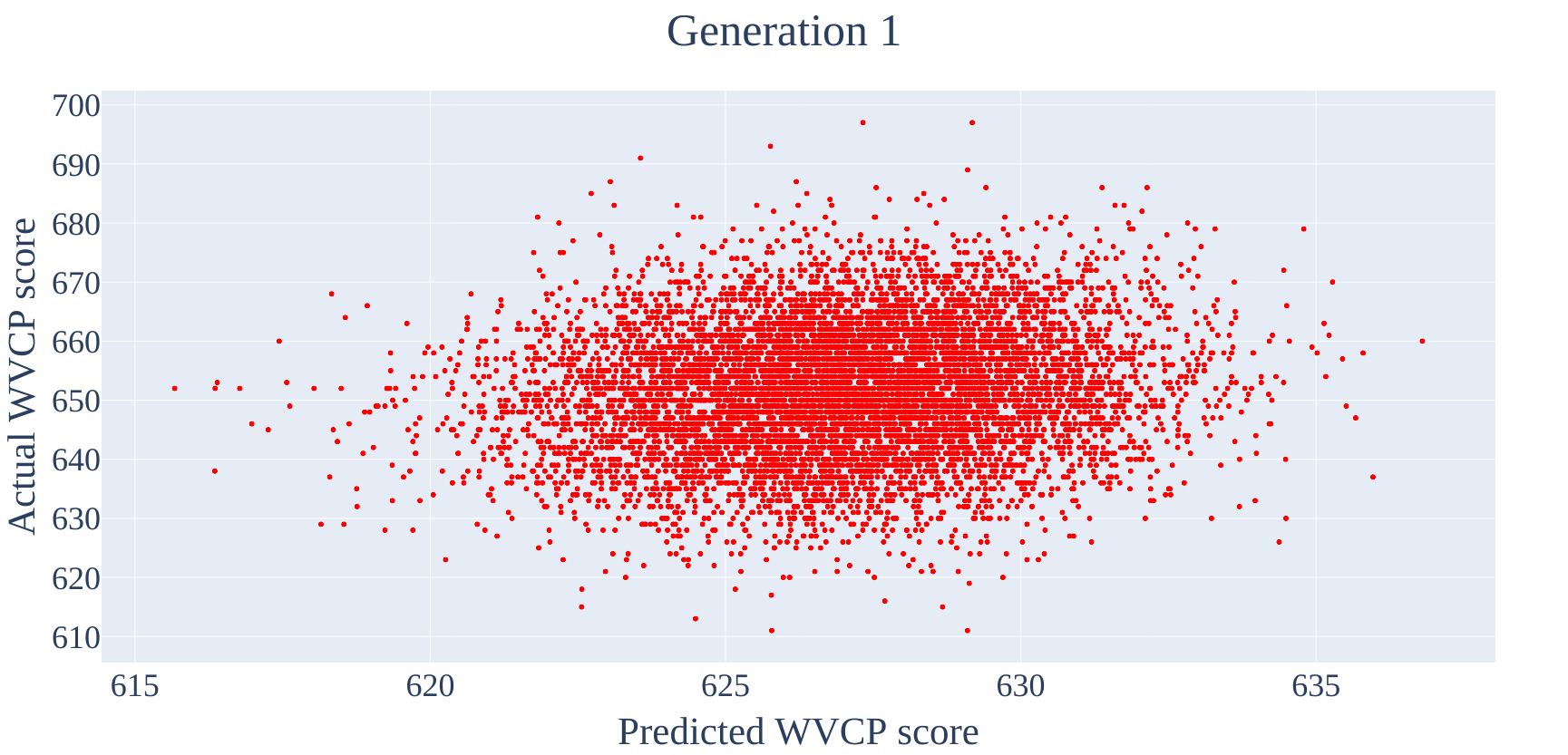}
    \includegraphics[width=0.45\textwidth]{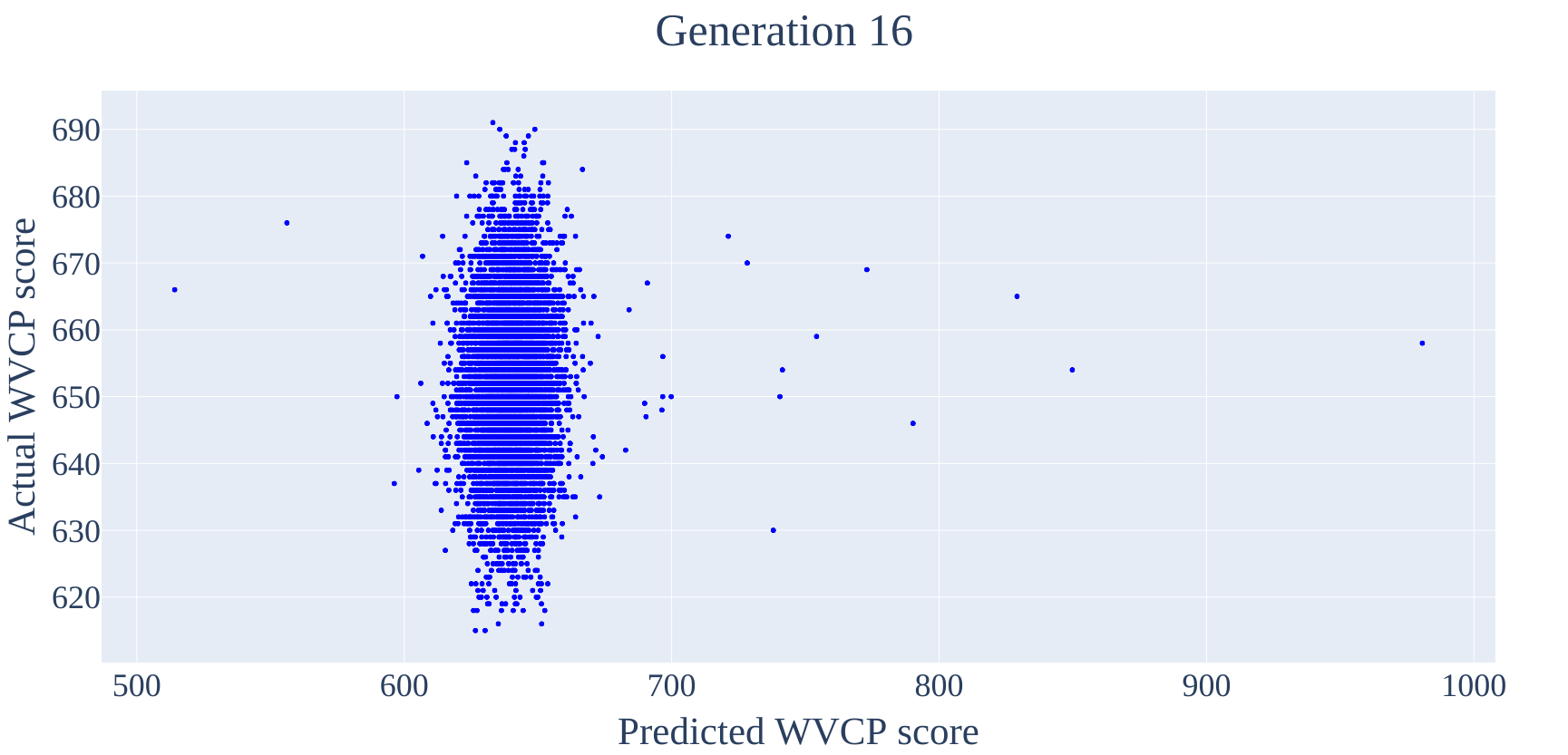}
    \includegraphics[width=0.45\textwidth]{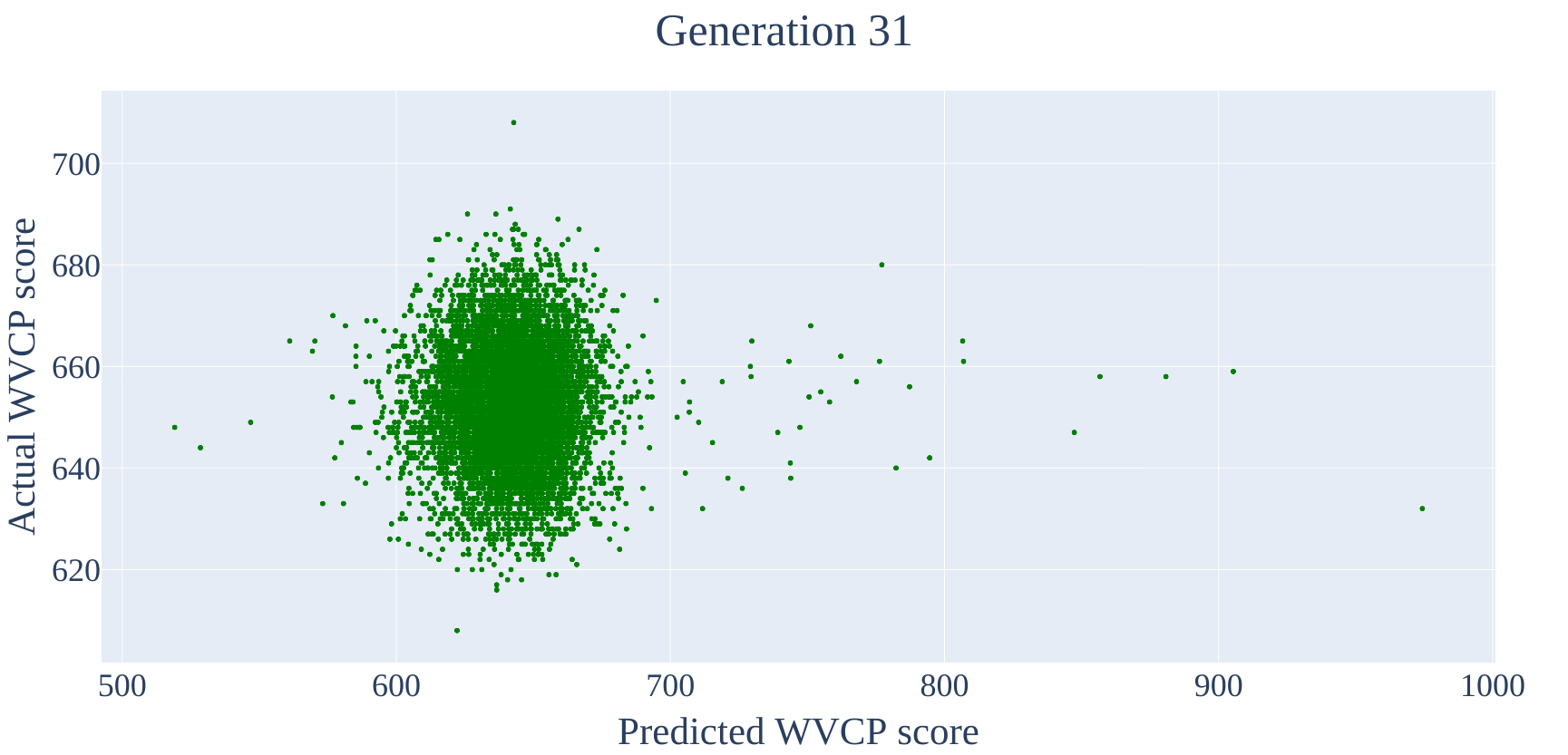}     
    \includegraphics[width=0.45\textwidth]{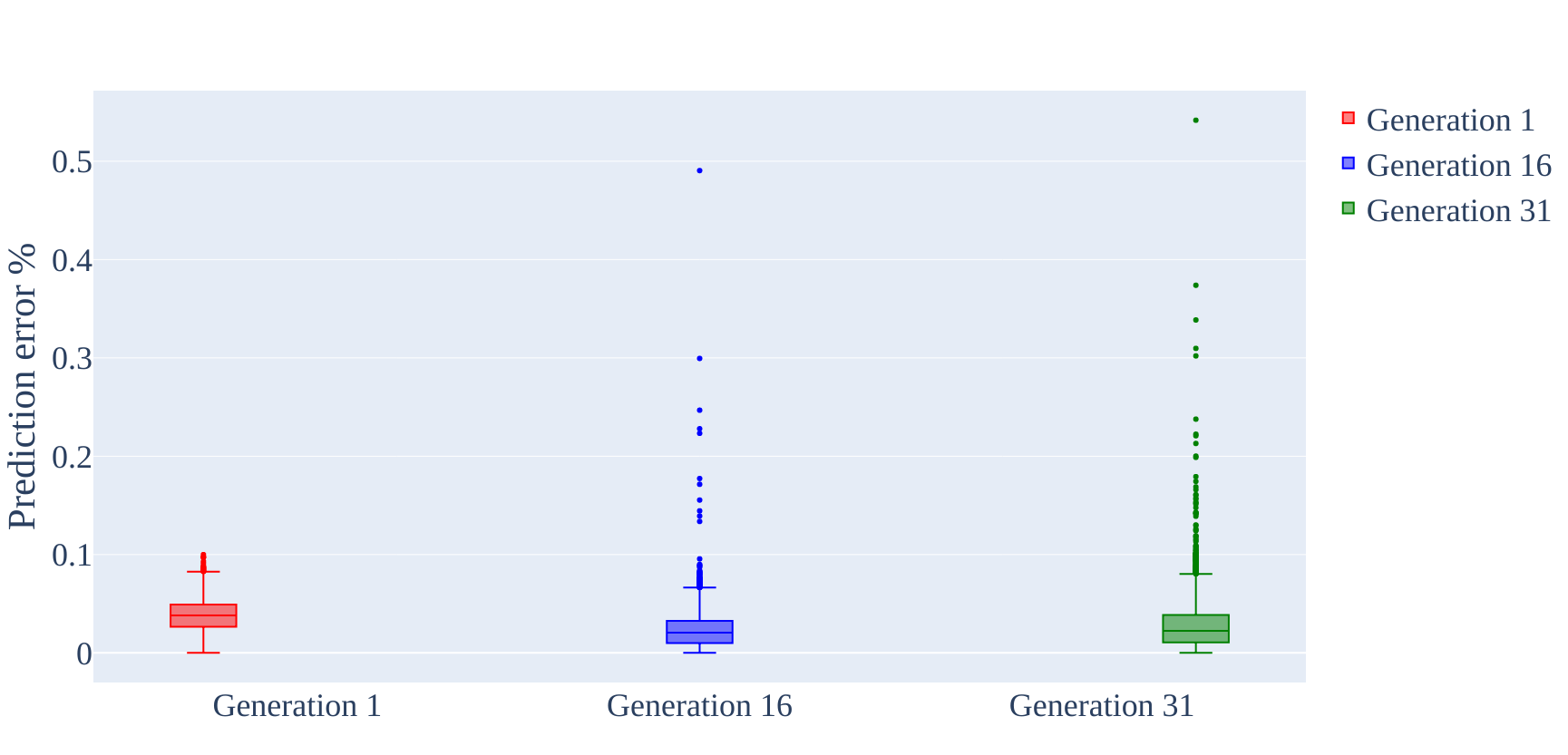} 
    \caption{Comparison of the predicted and actual results of the local searches at generation 1, 16 and 31 for the instance wap05a.col.}
    \label{fig:expected_results2}
\end{figure*}

We recorded these predicted and actual fitness values at every generation of the search process for several instances of the WVCP (DSJC500.5.col, wap05a.col, DSJC1000.1.col, le450\_25c.col and le450\_25d.col). In Figures \ref{fig:expected_results1}, and \ref{fig:expected_results2}, we  present two typical patterns of the evolution of the quality of the neural network prediction over the generations that we observed for these instances. For some graphs such as DSJC500.5.col, the neural network make more and more precise predictions on average over generations, but for other graphs such as wap05a.col, the neural network does not really improve its predictions over time.

Figure \ref{fig:expected_results1} displays three scatter plots at generation 1, 16 and 31 where, the $x$-axis and  $y$-axis respectively correspond to the predicted WVCP scores (generation 0, 15 and 30) and the actual WVCP scores (generation 1, 16 and 31) obtained after the local search for the instance DSJC500.5.col for all the $p = 20480$ individuals of the whole population. In the bottom right corner is displayed a boxplot of the prediction error in percent for the $p = 20480$ local searches at generation 1, 16 and 31.
 One first observes that the neural network is quite inaccurate at generation 1 (in red), because the relative prediction error is quite high, around 8.7\% and the Pearson correlation coefficient between the predicted and actual results is equal to $0.015$.  However, at generation 16 (in blue), the neural network can  provide more accurate predictions  of the  WVCP score that can be obtained by the local searches.  Indeed the relative error is below 1.4\% and the Pearson correlation coefficient gets to the value of $0.42$.  At generation 31, we observe that the neural network makes huge mistakes in the prediction, but only for one part of the samples, which can be explained by the fact that the prediction of the correct fitness values  becomes more and more difficult as the search for very good solutions becomes more unpredictable. However at generation 31, the median of the relative error is lower for the whole population.
 
Figure \ref{fig:expected_results2} displays the same comparison between expected and actual results of the local searches for the large graph wap05a.col with a low edge density for which we have shown in Section \ref{sec:benchmarks} that DLMCOL fails for this type of instance. An attempt to explain this failure can be seen in Figure \ref{fig:expected_results2}, where we observe that the relative prediction error is always high (around 2 and 3 \%), but more importantly, the Pearson correlation calculated between the predicted and actual scores is respectively equal to 0.13, 0.05 and 0.01 at generation 1, 16 and 31, which means that the neural network is not able to really distinguish promising new starting points for local search among all the possible ones.

It can be explained by the fact that good solutions for this type of instances are typically characterized by a low ratio of the number of color groups over the total number of vertices. As for the WVCP, only the maximum weight of each color group has an impact on the score, many different groupings of vertices are possible without impacting the score for these instances. It results in a very high average distance between the best solutions of the population. Therefore, the neural network fails to learn relevant patterns in this too diversified population.
  
\begin{figure*}[!h]
    \centering
    \includegraphics[width=0.8\textwidth]{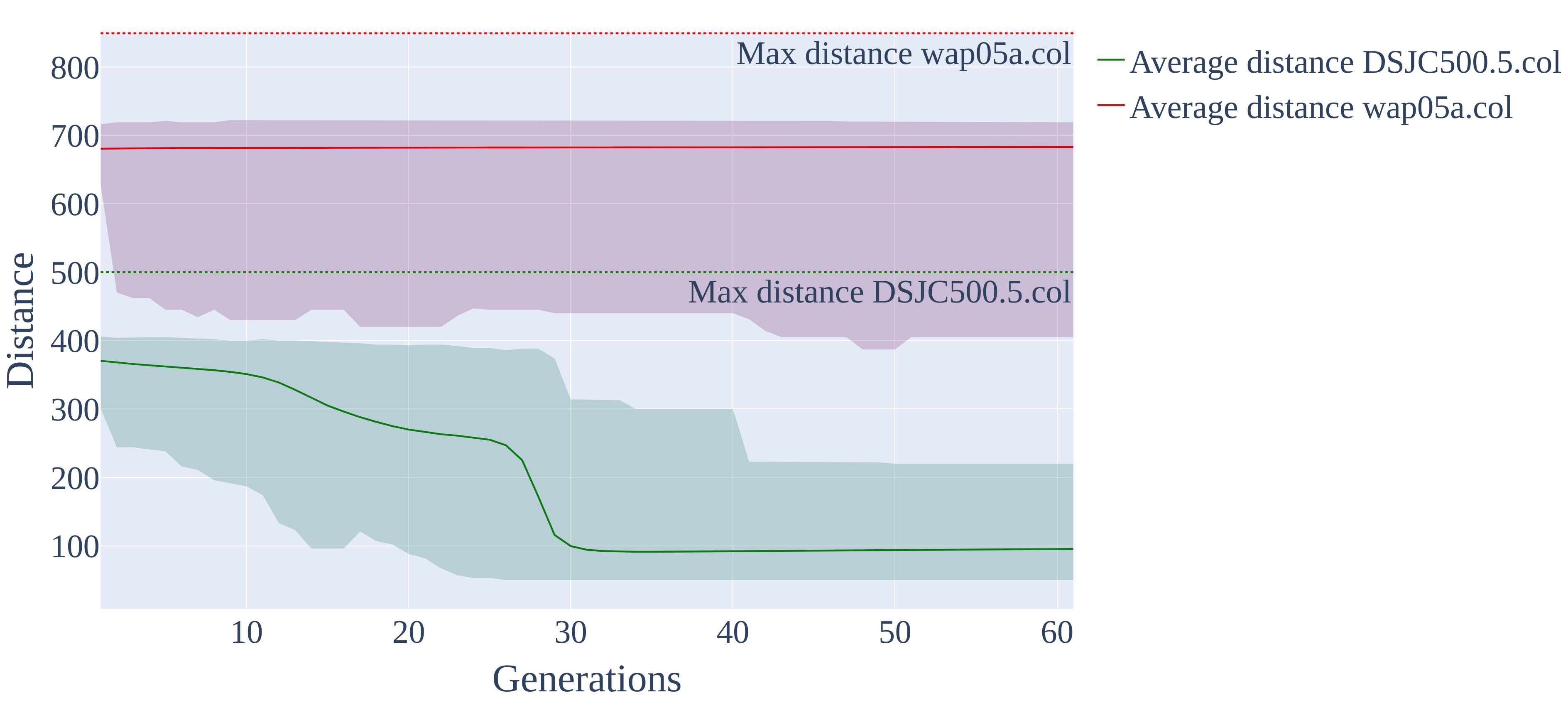}
    \caption{Red and green solid lines correspond to the average of the distance between the individuals of the population for the instances wap05a.col and DSJC500.5.col. Green and red areas are delimited by the maximum and minimum values of the distances between all  the individuals in the population. The distance is an approximation of the set-theoretic partition distance (see Section \ref{DistanceComputation}).}
    \label{fig:distance_pop}
\end{figure*}

Figure \ref{fig:distance_pop} confirms this intuition by showing that the average distance between the individuals in the population (red solid line) remains very high over the generations (around 680 for a maximum of 849) during the resolution of the instance wap05a.col. 
   Conversely for the instance DSJC500.5.col,  the average distance between individuals (green solid line) decreases over generations, showing that the best individuals retained in the population share  backbones of good solutions. 
   
   The green and red areas in Figure \ref{fig:distance_pop} are delimited by the maximum and minimum values of the distances between all  the individuals in the population for the instances   wap05a.col (red) and DSJC500.5.col (green). We remark that a minimum distance of 50 is reach after 25 generations for the instance DSJC500.5.col and this distance does not drop below 50. It comes from the fact that a minimum spacing between the individuals is imposed during the insertion of new individuals in the population (cf. Section. \ref{PopulationUpdate}).
    
\section{Conclusion}

A deep learning guided memetic framework for graph coloring problems was presented, as well as an implementation on GPU devices to solve the classical vertex $k$-coloring problem and the weighted vertex coloring problem. This approach uses the deep set architecture to learn an invariant by color permutation  regression model, useful to select the most promising crossovers at each generation. Additionally, it can take advantage of GPU computations to perform massively parallel local optimization with a large population to ensure a high degree of search intensification while maintaining a suitable degree of population diversification.

The proposed approach was assessed on popular DIMACS and COLOR challenge benchmarks of the two studied coloring problems. The computational results show that the algorithm competes globally well with the best algorithms for both problems. For the vertex coloring problem, it can reach most of the best known results of the literature for difficult instances. For the weighted coloring problem, it can find 14 new upper bounds for very difficult instances and significantly improves the previous best results for three graphs. An analysis of the predicted and actual fitness values after local search shows that the neural network can help to some extent in finding  promising new good starting points at each generation, which eases the discovery of high quality solutions in the search space. 

The achieved results reveal however three main limitations of the proposed approach. First, due to the memory capacity on the GPU devices we used, the DLMCOL algorithm has trouble to deal with very large instances ($|V| \geq 1000$). In particular, for the parallel local searches, the memory available on each thread of the GPU can be a huge limitation. Secondly, the algorithm has a slow convergence in comparison with sequential local search algorithms, due to its large population and the time spent to train the neural network at each generation. Thirdly, the algorithm fails for large instances with a low density (sparse graphs) for the WVCP, as for these instances the neural network has trouble to learn good patterns to effectively guide the selection of promising crossovers.

Other future works could be envisaged. In particular, it would be interesting to test the DLMCOL framework with the same type of neural network architecture to solve other graph coloring problems. Moreover, it could be worth applying deep learning techniques to learn a specific crossover for the weighted graph coloring problem instead of the classical GPX crossover used in this work. Finally, other neural network structures, such as graph convolutional neural networks, could be investigated to overcome the difficulty encountered on sparse graphs.

\section*{Declaration of competing interest}
The authors declare that they have no known competing financial interests or personal relationships that could have appeared to influence the work reported in this paper.

\section*{Acknowledgment}

We would like to thank the reviewers for their useful comments. We also thank Dr. Wen Sun for sharing the binary code of the AFISA algorithm \cite{sun2017feasible}, Dr. Yiyuan Wang for sharing the source code of the RedLS algorithm \cite{wang2020reduction} and Pr. Bruno Nogueira for sharing the source code of the ILS-TS algorithm \cite{nogueira2021iterated}. This work was granted access to the HPC resources of IDRIS (Grant No. 2020-A0090611887) from GENCI, which is also acknowledged.

\bibliography{biblio}
\bibliographystyle{plain}

\newpage

\appendix

\section{Results of DLMCOL on easy instances of vertex coloring problem \label{app:easyGCP}}

This appendix (Table \ref{table:gcp_easy}) reports the computational results reached by the DLMCOL algorithm on the easy set of the DIMACS challenge benchmark graphs. As Table \ref{table:gcp_easy} shows, DLMCOL can consistently and easily reach the chromatic number $\chi(G)$ or the best known result $k^*$ for each instance.

\begin{table}[!h]
\centering
\scriptsize
\caption{Computational results of DLMCOL on the \textit{easy} DIMACS challenge benchmarks for the COL problem}.

\begin{tabular}{l|ll|ll|lll} 
   \hline
   & \multicolumn{4}{c}{}  &  \multicolumn{3}{|c}{DLMCOL}\\
   \hline
  Instance & $|V|$ & dens. & $\chi(G)$ & $k^*$ & $k_{\text{best}}$ & SR & t(s) \\
    \hline
    DSJC125.1 & 125 & 0.1 & 5 & 5   &  \textbf{5} & 10/10  & 19\\
    DSJC125.5 & 125 & 0.5 & 17 & 17   &  \textbf{17} & 10/10  & 27\\
    DSJC125.9 & 125 & 0.5 & 44 & 44   &  \textbf{44} & 10/10  & 33\\
    DSJC250.1 & 250 & 0.1 & ? & 8   &  \textbf{8} & 10/10  & 36\\
    DSJC250.9 & 250 & 0.9 & 72 & 72   &  \textbf{72} & 10/10  & 313\\
    r125.1 & 125 & 0.03 & 5 & 5   &  \textbf{5} & 10/10  & 15\\
    r125.1c & 125 & 0.97 & 46 & 46   &  \textbf{46} & 10/10  & 32\\
    r125.5 & 125 & 0.5 & 36 & 36   &  \textbf{36} & 10/10  & 322\\
    r250.1 & 250 & 0.03 & 8 & 8   &  \textbf{8} & 10/10  & 23\\
    r250.1c & 250 & 0.97 & 64 & 64   &  \textbf{64} & 10/10  & 73\\
   DSJR500.1 & 500 & 0.03 & 12 & 12   &  \textbf{12} & 10/10  & 64\\
   r1000.1 & 1000 & 0.03 & 20 & 20   &  \textbf{20} & 10/10  & 258\\
   le450\_5a & 450 & 0.06 & 5 & 5   &  \textbf{5} & 10/10  & 69\\
   le450\_5b & 450 & 0.06 & 5 & 5   &  \textbf{5} & 10/10  & 76\\
   le450\_5c & 450 & 0.10 & 5 & 5   &  \textbf{5} & 10/10  & 76\\
   le450\_5d & 450 & 0.10 & 5 & 5   &  \textbf{5} & 10/10  & 69\\
   le450\_15a & 450 & 0.08 & 15 & 15   &  \textbf{15} & 10/10  & 95\\
   le450\_15b & 450 & 0.08 & 15 & 15   &  \textbf{15} & 10/10  & 93\\
   le450\_15c & 450 & 0.17 & 15 & 15   &  \textbf{15} & 10/10  & 143\\
   le450\_15d & 450 & 0.17 & 15 & 15   &  \textbf{15} & 10/10  & 314\\
   le450\_25a & 450 & 0.08 & 25 & 25   &  \textbf{25} & 10/10  & 58\\
   le450\_25b & 450 & 0.08 & 25 & 25   &  \textbf{25} & 10/10  & 55\\
   school1 & 385 & 0.26 & 14 & 14   &  \textbf{14} & 10/10  & 73\\
   school1\_nsh & 352 & 0.24 & 14 & 14   &  \textbf{14} & 10/10  & 59\\
   flat300\_20\_0 & 300 & 0.48 & 20 & 20   &  \textbf{20} & 10/10  & 56\\
    \hline
\end{tabular}
\label{table:gcp_easy}
\end{table}

\end{document}